\documentclass[11pt]{article}

\usepackage[utf8]{inputenc}
\usepackage[T1]{fontenc}
\usepackage{lmodern}
\usepackage[margin=1in]{geometry}
\usepackage[numbers,sort&compress]{natbib}
\usepackage{hyperref}
\usepackage{url}
\usepackage{booktabs}
\usepackage{amsfonts}
\usepackage{nicefrac}
\usepackage{microtype}
\usepackage{xcolor}
\usepackage{graphicx}
\usepackage{amsmath}
\usepackage{amssymb}
\usepackage{multirow}
\usepackage{subcaption}

\title{Collaborative Parameter Learning: Mitigating Forgetting via Parameter-Level Gradient Analysis}
\author{
\begin{tabular}{c}
\small Mutian Yang\textsuperscript{1,*}\quad
Zisen Zhan\textsuperscript{2,*}\quad
Yutong Chen\textsuperscript{1}\quad
Haolin Li\textsuperscript{1}\quad
Kaiwen Wang\textsuperscript{1}\\
\small Kaili Zheng\textsuperscript{1}\quad
Yuguang Wang\textsuperscript{2}\quad
Qi Wang\textsuperscript{3,\textdagger}\quad
Jiandong Gao\textsuperscript{1,\textdagger}\quad
Ji Wu\textsuperscript{1,4,5,\textdagger}\\[0.8em]
\footnotesize \textsuperscript{1}Department of Electronic Engineering, Tsinghua University, Beijing, China\\
\footnotesize \textsuperscript{2}Institute of Medical Technology, Peking University Health Science Center,\\
\footnotesize Peking University, Beijing, China\\
\footnotesize \textsuperscript{3}College of Information Science and Engineering, Northeastern University, Shenyang, China\\
\footnotesize \textsuperscript{4}College of AI, Tsinghua University, Beijing, China\\
\footnotesize \textsuperscript{5}Beijing National Research Center for Information Science and Technology, Beijing, China\\[0.8em]
\footnotesize\ttfamily yangmutian@mail.tsinghua.edu.cn\quad zisen\_zhan\_25@stu.pku.edu.cn\\
\footnotesize\ttfamily wangqi@ise.neu.edu.cn\quad jdgao@tsinghua.edu.cn\quad wuji\_ee@tsinghua.edu.cn\\[0.5em]
\footnotesize \textsuperscript{*}Equal contribution.\quad \textsuperscript{\textdagger}Corresponding author.
\end{tabular}
}
\date{}

\begin{document}

\maketitle

\begin{abstract}
Catastrophic forgetting during knowledge injection impairs the ability of large language models to acquire new knowledge without overwriting previously mastered knowledge. Recent studies analyze forgetting from a gradient-similarity perspective and mitigate forgetting through vector projection. However, these methods primarily characterize gradient similarity at the aggregate direction level, leaving the parameter-wise contributions to forgetting underexplored. In this paper, we decompose gradient similarity into parameter-wise contributions and identify two types of parameters during forgetting: \textbf{Conflicting Parameters}, whose updates contribute to forgetting and typically account for 50\%–75\% of parameters, and \textbf{Collaborative Parameters}, whose updates mitigate forgetting and account for 25\%–50\%. Based on this analysis, we propose \textbf{Collaborative Parameter Learning} (CPL), a parameter-wise training rule that freezes Conflicting Parameters and updates only Collaborative Parameters. Experiments comparing CPL with seven baseline methods show that CPL learns 20.2\%–48.2\% more questions with negligible forgetting, while reducing peak VRAM by approximately 3 GB per billion model parameters and computation time by 16.5\%. Extensive evaluations on parameter consumption, out-of-set generalization, cross-prompt generalization, multimodal tasks, open-ended question answering, and multilingual settings demonstrate that CPL effectively mitigates forgetting across diverse scenarios.
\end{abstract}

\section{Introduction}

Large language models (LLMs) have shown exceptional knowledge capacity and achieved remarkable potential across a broad range of domains \cite{achiam2023gpt,touvron2023llama}. To expand their knowledge breadth, various knowledge injection methods have been proposed \cite{lewis2020rag,wang2024knowledge,ke2023continual}. Although these methods enable LLMs to acquire new knowledge, they also cause the forgetting of previously mastered knowledge \cite{zhouinvestigating}. This phenomenon, known as catastrophic forgetting \cite{korbak2022controlling}, severely undermines the preservation of previously acquired knowledge in LLMs \cite{zhou2024continual}.

Existing literatures on mitigating catastrophic forgetting are typically categorized into four primary paradigms \cite{de2021continual}. Regularization-based methods penalize changes to parameters critical for previously mastered knowledge \cite{kirkpatrick2017overcoming,aljundi2018memory}, while parameter-isolation techniques alleviate interference by confining updates to dedicated modules or parameter subsets \cite{wang2021kadapter,hu2022lora}. Among these paradigms, gradient-based methods are particularly significant because they formalize forgetting as an optimization-level conflict between injected and mastered knowledge \cite{yu2020gradient,liu2021conflict}. By characterizing interference through gradient similarity, these methods mitigate forgetting by projecting the update direction to reduce detrimental interference with mastered knowledge \cite{farajtabar2020orthogonal,chaudhry2019efficient,saha2021gradient}.

However, most existing gradient-based methods treat gradient interference as a single, aggregate vector-level quantity \cite{farajtabar2020orthogonal,chaudhry2019efficient,saha2021gradient}, enforcing strict orthogonality between newly injected and previously mastered knowledge to mitigate forgetting. As a result, they overlook the parameter-wise contributions to forgetting and impose overly restrictive constraints on knowledge injection.

Therefore, critical questions arise: do all parameters contribute to forgetting, or only a subset of them? How to eliminate the conflicting components between injected and mastered knowledge while preserving the collaborative components, rather than strictly enforcing orthogonality? This work attempts to answer these questions as follows:

$\bullet$ How to characterize catastrophic forgetting from the perspective of gradients? In Section 3, we analyze the relationship between gradient similarity and forgetting, revealing that samples with strongly negative gradient similarity are more susceptible to forgetting.

$\bullet$ Which parameters are responsible for catastrophic forgetting? In Section 4, we reveal two distinct types of parameters: \textbf{Conflicting Parameters} and \textbf{Collaborative Parameters}. Conflicting Parameters induce forgetting and typically comprise 50\%–75\% of all parameters, whereas Collaborative Parameters alleviate forgetting and constitute 25\%–50\%.

$\bullet$ How to mitigate catastrophic forgetting? In Section 5, we propose a novel knowledge injection method, \textbf{Collaborative Parameter Learning} (CPL), which freezes Conflicting Parameters and trains Collaborative Parameters. By preserving the collaborative directions between injected and mastered knowledge, rather than enforcing strict orthogonality, CPL learns 20.2\%–48.2\% more questions with negligible forgetting. By replacing high-cost gradient projection with low-cost sign comparison, CPL reduces peak VRAM by 3 GB per billion model parameters and computation time by 16.5\%.

$\bullet$ Is CPL robustness to diverse scenarios? In Section 6, we extend CPL to out-of-set generalization, cross-prompt transfer, multimodal tasks, open-ended question answering, and multilingual settings. Results indicate that CPL effectively mitigates forgetting in those settings.

The appendix provides additional experimental details and results.

\begin{figure*}[!t] 
    \centering
    \includegraphics[width=1\textwidth]{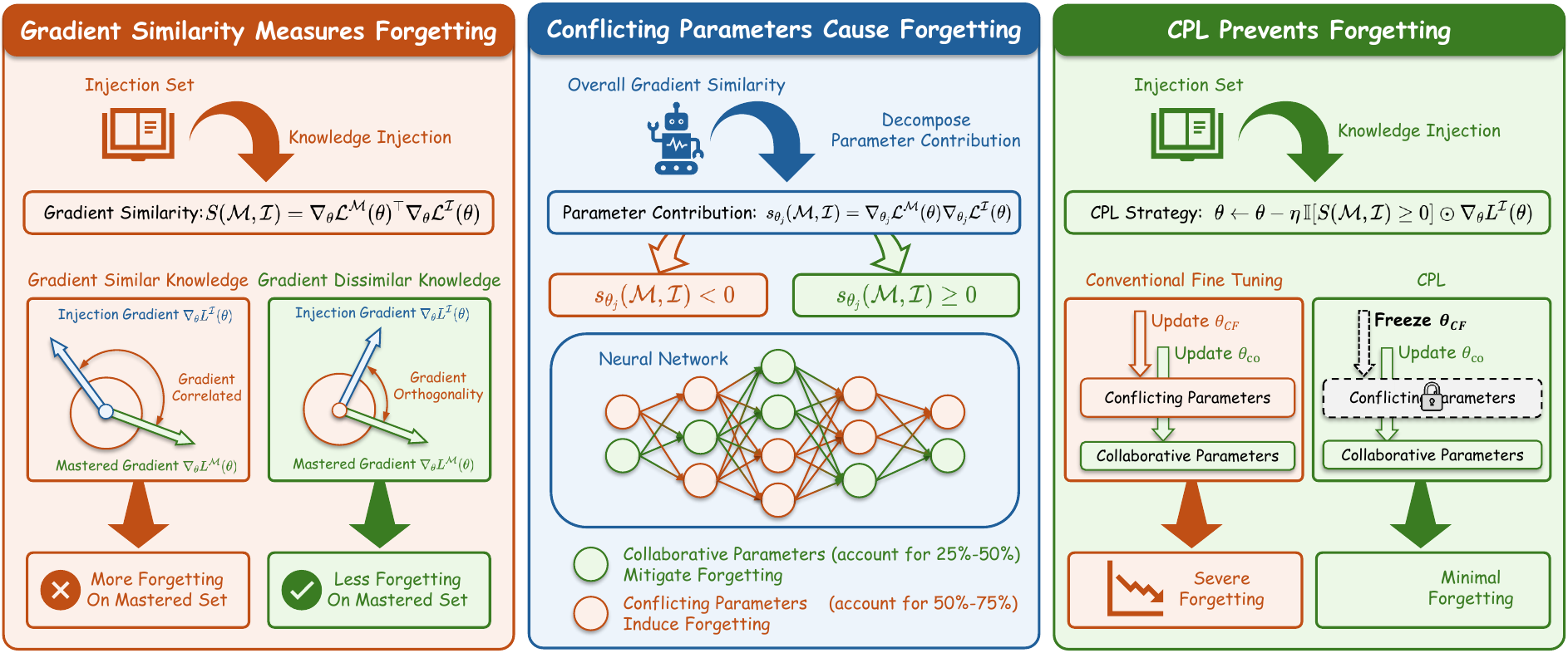} 
    \caption{Conceptual illustration of our framework. We decompose gradient similarity into parameter-wise contributions, and propose a parameter-wise training rule to mitigate forgetting.}
    \label{fig:conceptual_framework}
\end{figure*}

\section{Definition of Learning and Forgetting}

\subsection{Learning}

The parameters of LLMs are denoted as $\theta = (\theta_1, \dots, \theta_{|\theta|}) \in \mathbb{R}^{|\theta|}$, where $\theta_j$ represents the $j$-th parameter. Let $\mathcal{I}$ denote a set of knowledge samples to be injected
\begin{equation}
    \mathcal{I} = \left\{(x_1^\mathcal{I}, y_1^\mathcal{I}), \dots, (x_{|\mathcal{I}|}^\mathcal{I}, y_{|\mathcal{I}|}^\mathcal{I})\right\}
\end{equation}
Loss on a single injection sample $(x, y) \in \mathcal{I}$ is expressed as $L^{(x,y)}(\theta) = \ell(f_\theta(x), y)$, where $\ell(\cdot, \cdot)$ denotes the loss function. The total loss on Injection Set $\mathcal{I}$ is expressed as
\begin{align}
    \mathcal{L}^\mathcal{I}(\theta) &= \frac{1}{|\mathcal{I}|} \sum_{(x,y) \in \mathcal{I}} L^{(x,y)}(\theta) \notag \\
    &= \frac{1}{|\mathcal{I}|} \sum_{(x,y) \in \mathcal{I}} \ell(f_\theta(x), y)
\end{align}
Let $\nabla_\theta \mathcal{L}^\mathcal{I}(\theta) \in \mathbb{R}^{|\theta| \times 1}$ denote the gradient of $\mathcal{L}^\mathcal{I}(\theta)$ with respect to $\theta$. Under gradient descent with a learning rate $\eta$, the parameter update is given by
\begin{align}
    \theta \leftarrow \theta - \eta \nabla_\theta \mathcal{L}^\mathcal{I}(\theta)
\label{eq:update_theta}
\end{align}
The loss after parameter update is approximated using a first-order Taylor expansion
\begin{align}
    &\mathcal{L}^\mathcal{I}\left(\theta - \eta \nabla_\theta \mathcal{L}^\mathcal{I}(\theta)\right) \notag \\
    &= \frac{1}{|\mathcal{I}|} \sum_{(x,y) \in \mathcal{I}} [L^{(x,y)} \left( \theta - \eta \nabla_\theta \mathcal{L}^\mathcal{I}(\theta) \right)] \notag \\
    &\approx \frac{1}{|\mathcal{I}|} \sum_{(x,y) \in \mathcal{I}} \left[ L^{(x,y)}(\theta) - \eta \nabla_\theta L^{(x,y)}(\theta)^\top \nabla_\theta \mathcal{L}^\mathcal{I}(\theta)\right] \notag \\
    &= \mathcal{L}^\mathcal{I}(\theta) - \eta \nabla_\theta \mathcal{L}^\mathcal{I}(\theta)^\top \nabla_\theta \mathcal{L}^\mathcal{I}(\theta)
\end{align}
Therefore, the change in loss is:
\begin{align}
    \Delta \mathcal{L}^\mathcal{I} &= \mathcal{L}^\mathcal{I}\left(\theta - \eta \nabla_\theta \mathcal{L}^\mathcal{I}(\theta)\right) - \mathcal{L}^\mathcal{I}(\theta) \notag \\
    &= - \eta \nabla_\theta \mathcal{L}^\mathcal{I}(\theta)^\top \nabla_\theta \mathcal{L}^\mathcal{I}(\theta)
\end{align}
Since $\nabla_\theta \mathcal{L}^\mathcal{I}(\theta)$ is column vector, $\nabla_\theta \mathcal{I}^\mathcal{I}(\theta)^\top \nabla_\theta \mathcal{L}^\mathcal{I}(\theta)$ represents the inner product of the gradient with itself. We define the \textbf{Gradient Similarity} as
\begin{equation}
    S(\mathcal{I}, \mathcal{I}) = \nabla_\theta \mathcal{L}^\mathcal{I}(\theta)^\top \nabla_\theta \mathcal{L}^\mathcal{I}(\theta)
\end{equation}
Due to the non-negativity of the inner product, we have $\Delta \mathcal{L}^\mathcal{I} = - \eta S(\mathcal{I}, \mathcal{I}) \leq 0$. This derivation characterizes knowledge injection: under a sufficiently small learning rate, updating along the negative gradient of $\mathcal{I}$ decreases its loss in the first-order approximation.

\subsection{Forgetting}

Subsequently, we analyze the impact of knowledge injection on samples with mastered knowledge, denoted as Mastered Set 
\begin{equation}
    \mathcal{M} = \left\{(x_1^\mathcal{M}, y_1^\mathcal{M}), \dots, (x_{|\mathcal{M}|}^\mathcal{M}, y_{|\mathcal{M}|}^\mathcal{M})\right\}
\end{equation}
The loss on $\mathcal{M}$ is defined as $\mathcal{L}^\mathcal{M}(\theta)$. After parameter update , the loss becomes $\mathcal{L}^\mathcal{M}\left(\theta - \eta \nabla_\theta \mathcal{L}^\mathcal{I}(\theta)\right)$. The expression is expanded through first-order Taylor approximation:
\begin{align}
    \label{eq:loss_exp}
    &\mathcal{L}^\mathcal{M}\left(\theta - \eta \nabla_\theta \mathcal{L}^\mathcal{I}(\theta)\right) \notag \\
    &= \frac{1}{|\mathcal{M}|} \sum_{(x,y) \in \mathcal{M}} L^{(x,y)} \left( \theta - \eta \nabla_\theta \mathcal{L}^\mathcal{I}(\theta) \right) \notag \\
    &\approx \frac{1}{|\mathcal{M}|} \sum_{(x,y) \in \mathcal{M}} \left[ L^{(x,y)}(\theta) - \eta \nabla_\theta L^{(x,y)}(\theta)^\top \nabla_\theta \mathcal{L}^\mathcal{I}(\theta)\right] \notag \\
    &= \mathcal{L}^\mathcal{M}(\theta) - \eta \nabla_\theta \mathcal{L}^\mathcal{M}(\theta)^\top \nabla_\theta \mathcal{L}^\mathcal{I}(\theta)
\end{align}
The change in loss is expressed as
\begin{align}
    \Delta \mathcal{L}^\mathcal{M} &= \mathcal{L}^\mathcal{M}\left(\theta - \eta \nabla_\theta \mathcal{L}^\mathcal{I}(\theta)\right) - \mathcal{L}^\mathcal{M}(\theta) \notag \\
    &= - \eta \nabla_\theta \mathcal{L}^\mathcal{M}(\theta)^\top \nabla_\theta \mathcal{L}^\mathcal{I}(\theta)
\end{align}
We define the gradient similarity between $\mathcal{M}$ and $\mathcal{I}$ as
\begin{equation}
    \label{eq:sim}
    S(\mathcal{M}, \mathcal{I}) = \nabla_\theta \mathcal{L}^\mathcal{M}(\theta)^\top \nabla_\theta \mathcal{L}^\mathcal{I}(\theta)
\end{equation}
Therefore, during knowledge injection, the loss change on $\mathcal{M}$ is given by
\begin{equation}
    \label{eq:delta_loss_E}
    \Delta \mathcal{L}^\mathcal{M} = - \eta S(\mathcal{M}, \mathcal{I})
\end{equation}
Eq.~\eqref{eq:delta_loss_E} reveals a critical insight: the loss on $\mathcal{M}$ increases when the gradient similarity is negative. Therefore, mastered samples with more negative gradient similarity (i.e., stronger alignment in the opposite direction) are more susceptible to catastrophic forgetting.

\section{Gradient Similarity Measures Catastrophic Forgetting}

\begin{table*}[t]
\centering
\caption{Relationship between gradient similarity and forgetting. Results are reported in the format of ``Number of forgotten questions (Percentage)''. Samples with negative gradient similarity are ranked by magnitude: the top $1/3$ with the largest magnitudes are labeled ``Sim'', while the bottom $1/3$ are labeled ``Dissim''. Results indicate that forgetting occurs primarily in the ``Sim'' group.}
\label{tab:gradient_similarity}
\begin{small}
\begin{sc}
\resizebox{\linewidth}{!}{
\begin{tabular}{lcccccccc}
\toprule
\multirow{2}{*}{\textbf{Model}} & \multicolumn{2}{c}{\textbf{MMLU}} & \multicolumn{2}{c}{\textbf{MedQA}} & \multicolumn{2}{c}{\textbf{ARC-C}} & \multicolumn{2}{c}{\textbf{CSQA}} \\
\cmidrule(lr){2-3} \cmidrule(lr){4-5} \cmidrule(lr){6-7} \cmidrule(lr){8-9}
 & \textbf{Dissim} & \textbf{Sim} & \textbf{Dissim} & \textbf{Sim} & \textbf{Dissim} & \textbf{Sim} & \textbf{Dissim} & \textbf{Sim} \\
\midrule
\multirow{2}{*}{\textit{Qwen 1.5B}} 
 & 4 & 137 & 37 & 113 & 3 & 72 & 10 & 102 \\
 & (1.4\%) & (48.9\%) & (20.6\%) & (62.8\%) & (1.0\%) & (23.2\%) & (3.7\%) & (38.1\%) \\
\midrule
\multirow{2}{*}{\textit{Qwen 3B}}    
 & 0 & 118 & 0 & 113 & 7 & 193 & 0 & 84 \\
 & (0.0\%) & (37.9\%) & (0.0\%) & (54.6\%) & (2.0\%) & (56.4\%) & (0.0\%) & (31.5\%) \\
\midrule
\multirow{2}{*}{\textit{Qwen 7B}}    
 & 3 & 120 & 3 & 116 & 0 & 17 & 0 & 40 \\
 & (0.9\%) & (36.0\%) & (1.2\%) & (46.0\%) & (0.0\%) & (5.1\%) & (0.0\%) & (13.7\%) \\
\midrule
\multirow{2}{*}{\textit{LLaMA 1B}}  
 & 0 & 68 & 0 & 43 & 6 & 91 & 9 & 89 \\
 & (0.0\%) & (58.6\%) & (0.0\%) & (42.2\%) & (6.6\%) & (100\%) & (10.1\%) & (100.0\%) \\
\midrule
\multirow{2}{*}{\textit{LLaMA 3B}}  
 & 0 & 83 & 28 & 32 & 1 & 73 & 0 & 34 \\
 & (0.0\%) & (33.2\%) & (11.1\%) & (12.6\%) & (0.4\%) & (25.9\%) & (0.0\%) & (13.6\%) \\
\bottomrule
\end{tabular}
}
\end{sc}
\end{small}
\vskip -0.1in
\end{table*}

\subsection{Experiment Procedure}

\textbf{Datasets:}
We employ standard benchmarks including Massive Multitask Language Understanding (MMLU) \cite{hendrycks2020measuring}, Medical Question Answering (MedQA) \cite{jin2021disease}, AI2 Reasoning Challenge (ARC-C) \cite{clark2018think}, Commonsense Question Answering (CSQA) \cite{talmor2019commonsenseqa}. Additionally, Yahoo Answers \cite{zhang2015character}, AG News \cite{zhang2015character}, and DBPedia \cite{zhang2015character}, NumGLUE-cm \cite{mishra2022numglue}, NumGLUE-ds \cite{mishra2022numglue}, C-STANCE \cite{zhao2023cstance}, and MMBench \cite{liu2024mmbench} are used for extensive study.

\textbf{Models:}
The experiments are conducted on \textit{Qwen} \cite{qwen2} and \textit{LLaMA} \cite{grattafiori2024llama} families, including \textit{Qwen2.5 1.5B}, \textit{Qwen2.5 3B}, \textit{Qwen2.5 7B}, \textit{LLaMA3.2 1B}, and \textit{LLaMA3.2 3B}. The multimodal evaluation is conducted on \textit{Qwen2.5vl 3B}.

\subsection{Catastrophic Forgetting Happens during Knowledge Injection}

To distinguish Injection Set $\mathcal{I}$ and Mastered Set $\mathcal{M}$, LLMs are asked to answer the questions in datasets (inference details in Appendix \ref{app:inference}). Correctly answered questions are grouped into $\mathcal{M}$, while incorrect ones form $\mathcal{I}$. Appendix \ref{app:inference_results} exhibits the distribution of $\mathcal{I}$ and $\mathcal{M}$.

LLMs are trained on $\mathcal{I}$ as detailed in Appendix ~\ref{app:training}. Samples in $\mathcal{I}$ are used to evaluate learning, while samples in $\mathcal{M}$ are used to measure forgetting. Table~\ref{tab:learning_forgetting_details} reveals that while LLMs successfully learn new knowledge from $\mathcal{I}$, they simultaneously suffer forgetting on $\mathcal{M}$. Similar forgetting is also observed in Low-Rank Adaptation (LoRA) setting \cite{hu2022lora} as shown in Appendix~\ref{app:lora}.

\subsection{Strongly Negative Gradient Similarity Leads to Forgetting}
\label{sec:gradient_similarity_analysis}

Gradient similarity between samples in $\mathcal{M}$ and $\mathcal{I}$ is computed. In light of Eq.~\eqref{eq:delta_loss_E}, samples with negative gradient similarity are ranked by the magnitude of their similarity. The top $1/3$ with the largest magnitudes are assigned to the ``Sim'' group, representing strongly negative similarity, while the bottom $1/3$ are assigned to the ``Dissim'' group.

Table~\ref{tab:gradient_similarity} indicates that forgetting is predominantly concentrated in the ``Sim'' group. This evidence provides support for Eq.~\eqref{eq:delta_loss_E}, confirming that strongly negative gradient similarity drives forgetting.

\section{Conflicting Parameters Cause Catastrophic Forgetting}
\label{sec:conflicting_neurons_cause_catastrophic_forgetting}

\begin{table}[t]
\centering
\caption{Distribution of Collaborative Parameters (COLL) and Conflicting Parameters (CONF). ``PROP'' represents the proportion of a specific parameter. ``GRAD'' represents the gradient similarity sum. ``TOTAL'' represents the gradient similarity sum for all parameters. Collaborative Parameters typically account for 25\%--50\%, while Conflicting Parameters account for 50\%--75\%.}
\label{tab:neuron_distribution}
\begin{small}
\begin{sc}
\resizebox{\linewidth}{!}{
\begin{tabular}{llcccccccc}
\toprule
\multirow{2}{*}{Model} & \multirow{2}{*}{Metric} & \multicolumn{2}{c}{MMLU} & \multicolumn{2}{c}{MedQA} & \multicolumn{2}{c}{ARC-C} & \multicolumn{2}{c}{CSQA} \\
\cmidrule(lr){3-4} \cmidrule(lr){5-6} \cmidrule(lr){7-8} \cmidrule(lr){9-10}
 & & Coll & Conf & Coll & Conf & Coll & Conf & Coll & Conf \\
\midrule
\multirow{3}{*}{\textit{Qwen 1.5B}} 
 & Prop  & 47.7\% & 52.3\% & 42.1\% & 57.9\% & 49.7\% & 50.3\% & 48.4\% & 51.6\% \\
 & Grad & +692   & $-$1851  & +567   & $-$3186  & +635    & $-$1021  & +577    & $-$1438  \\
 & Total  & \multicolumn{2}{c}{$-$1159} & \multicolumn{2}{c}{$-$2619} & \multicolumn{2}{c}{$-$386} & \multicolumn{2}{c}{$-$861} \\
\midrule
\multirow{3}{*}{\textit{Qwen 3B}}    
 & Prop  & 49.4\% & 50.6\% & 44.6\% & 55.4\% & 51.7\% & 48.3\% & 52.9\% & 47.1\% \\
 & Grad & +1626  & $-$2979  & +1538  & $-$4319  & +2237   & $-$2569  & +2703   & $-$3276  \\
 & Total  & \multicolumn{2}{c}{$-$1353} & \multicolumn{2}{c}{$-$2781} & \multicolumn{2}{c}{$-$332} & \multicolumn{2}{c}{$-$573} \\
\midrule
\multirow{3}{*}{\textit{Qwen 7B}}    
 & Prop  & 49.6\% & 50.4\% & 49.0\% & 51.0\% & 51.2\% & 48.8\% & 51.6\% & 48.4\% \\
 & Grad & +710   & $-$1242  & +661   & $-$1282  & +433    & $-$664   & +687    & $-$940   \\
 & Total  & \multicolumn{2}{c}{$-$532}  & \multicolumn{2}{c}{$-$621}  & \multicolumn{2}{c}{$-$231} & \multicolumn{2}{c}{$-$253} \\
\midrule
\multirow{3}{*}{\textit{LLaMA 1B}}  
 & Prop  & 34.1\% & 65.9\% & 24.0\% & 76.0\% & 28.6\% & 71.4\% & 42.1\% & 57.9\% \\
 & Grad & +2116  & $-$64263 & +54    & $-$116388& +373    & $-$88549 & +4349   & $-$41111 \\
 & Total  & \multicolumn{2}{c}{$-$62147} & \multicolumn{2}{c}{$-$116334}& \multicolumn{2}{c}{$-$88176}& \multicolumn{2}{c}{$-$36762} \\
\midrule
\multirow{3}{*}{\textit{LLaMA 3B}}  
 & Prop  & 37.8\% & 62.2\% & 30.4\% & 69.6\% & 46.5\% & 53.5\% & 42.7\% & 57.3\% \\
 & Grad & +282   & $-$2528  & +116   & $-$2784  & +734    & $-$2161  & +404    & $-$2178  \\
 & Total  & \multicolumn{2}{c}{$-$2246} & \multicolumn{2}{c}{$-$2668} & \multicolumn{2}{c}{$-$1427} & \multicolumn{2}{c}{$-$1774} \\
\bottomrule
\end{tabular}
}
\end{sc}
\end{small}
\vskip -0.1in
\end{table}

While LLMs exhibit overall forgetting on $\mathcal{M}$, it remains unclear whether forgetting is driven by all parameters or only a subset. To answer this question, gradient similarity of the $j$-th parameter between $\mathcal{M}$ and $\mathcal{I}$ is defined as
\begin{equation}
    s_{\theta_j}(\mathcal{M}, \mathcal{I}) = \nabla_{\theta_j} \mathcal{L}^\mathcal{M}(\theta) \nabla_{\theta_j} \mathcal{L}^\mathcal{I}(\theta)
\end{equation}
$\nabla_{\theta_j} \mathcal{L}^\mathcal{M}(\theta)$ and $\nabla_{\theta_j} \mathcal{L}^\mathcal{I}(\theta)$ represent the components of gradients on $\theta_j$ for $\mathcal{M}$ and $\mathcal{I}$, respectively. The global gradient similarity of LLMs is equivalent to the sum of gradient similarities of individual parameters. Therefore, Eq.~\eqref{eq:sim} is expressed as
\begin{align}
    S(\mathcal{M}, \mathcal{I}) &= \sum_{j=1}^{|\theta|} \left( \nabla_{\theta_j} \mathcal{L}^\mathcal{M}(\theta) \nabla_{\theta_j} \mathcal{L}^\mathcal{I}(\theta) \right) \notag \\
    &= \sum_{j=1}^{|\theta|} s_{\theta_j}(\mathcal{M}, \mathcal{I})
\end{align}
If the gradient similarity of $\theta_j$ with respect to $\mathcal{M}$ and $\mathcal{I}$ is negative, this parameter contributes to an increase in loss and induces forgetting. We refer to such parameters as \textbf{Conflicting Parameters}. Conversely, if the gradient similarity of $\theta_j$ is positive, this parameter contributes to a decrease in loss and mitigates forgetting. Such parameters are referred to as \textbf{Collaborative Parameters}.

Table~\ref{tab:neuron_distribution} describes the distribution of parameters. Approximately 25\%--50\% of parameters have positive gradient similarity and are termed Collaborative Parameters, while 50\%--75\% have negative contributions and are termed Conflicting Parameters. Moreover, the positive contributions from Collaborative Parameters are generally insufficient to offset the stronger negative contributions from Conflicting Parameters, causing the overall gradient similarity to remain negative.

\section{Mitigating Catastrophic Forgetting by Freezing Conflicting Parameters}

\subsection{Establishment of CPL}
\begin{table*}[t]
\centering
\caption{Performance of knowledge injection using FT and CPL. Results are reported in the format of ``Number of questions (Percentage)''. FT leads to significant forgetting of previously mastered knowledge, whereas CPL induces negligible catastrophic forgetting when $\mathcal{M}$ is fully available.}
\label{tab:ft_vs_cbl}
\begin{small}
\begin{sc}
\resizebox{\textwidth}{!}{
\begin{tabular}{llcccccccc}
\toprule
\multirow{2}{*}{\textbf{Model}} & \multirow{2}{*}{\textbf{Method}} & \multicolumn{2}{c}{\textbf{MMLU}} & \multicolumn{2}{c}{\textbf{MedQA}} & \multicolumn{2}{c}{\textbf{ARC-C}} & \multicolumn{2}{c}{\textbf{CSQA}} \\
\cmidrule(lr){3-4} \cmidrule(lr){5-6} \cmidrule(lr){7-8} \cmidrule(lr){9-10}
 & & \textbf{Learned} & \textbf{Forgot} & \textbf{Learned} & \textbf{Forgot} & \textbf{Learned} & \textbf{Forgot} & \textbf{Learned} & \textbf{Forgot} \\
\midrule
\multirow{4}{*}{\textit{Qwen 1.5B}} 
 & \multirow{2}{*}{FT}  & 168 & 188 & 191 & 264 & 104 & 88 & 86 & 163 \\
 &                      & (24.8\%) & (21.4\%) & (27.1\%) & (46.5\%) & (32.1\%) & (9.3\%) & (31.6\%) & (19.8\%) \\
\cmidrule(lr){2-10}
 & \multirow{2}{*}{CPL} & 168 & \textbf{0} & 193 & \textbf{0} & 163 & \textbf{0} & 147 & \textbf{0} \\
 &                      & (24.8\%) & (\textbf{0.0\%}) & (27.4\%) & (\textbf{0.0\%}) & (50.3\%) & (\textbf{0.0\%}) & (54.0\%) & (\textbf{0.0\%}) \\
\midrule
\multirow{4}{*}{\textit{Qwen 3B}} 
 & \multirow{2}{*}{FT}  & 114 & 126 & 119 & 160 & 71 & 286 & 77 & 105 \\
 &                      & (19.5\%) & (13.0\%) & (18.6\%) & (25.3\%) & (33.5\%) & (27.0\%) & (31.7\%) & (12.3\%) \\
\cmidrule(lr){2-10}
 & \multirow{2}{*}{CPL} & 116 & \textbf{0} & 119 & \textbf{0} & 76 & \textbf{0} & 81 & \textbf{0} \\
 &                      & (19.8\%) & (\textbf{0.0\%}) & (18.6\%) & (\textbf{0.0\%}) & (35.8\%) & (\textbf{0.0\%}) & (33.3\%) & (\textbf{0.0\%}) \\
\midrule
\multirow{4}{*}{\textit{Qwen 7B}} 
 & \multirow{2}{*}{FT}  & 127 & 193 & 110 & 169 & 62 & 19 & 68 & 48 \\
 &                      & (28.0\%) & (17.5\%) & (21.9\%) & (21.9\%) & (46.3\%) & (1.7\%) & (34.5\%) & (5.3\%) \\
\cmidrule(lr){2-10}
 & \multirow{2}{*}{CPL} & 189 & \textbf{0} & 249 & \textbf{0} & 90 & \textbf{0} & 144 & \textbf{0} \\
 &                      & (41.6\%) & (\textbf{0.0\%}) & (49.6\%) & (\textbf{0.0\%}) & (67.2\%) & (\textbf{0.0\%}) & (73.1\%) & (\textbf{0.0\%}) \\
\midrule
\multirow{4}{*}{\textit{LLaMA 1B}} 
 & \multirow{2}{*}{FT}  & 158 & 73 & 105 & 43 & 328 & 154 & 333 & 176 \\
 &                      & (14.0\%) & (17.0\%) & (11.0\%) & (13.4\%) & (34.4\%) & (48.7\%) & (47.8\%) & (44.1\%) \\
\cmidrule(lr){2-10}
 & \multirow{2}{*}{CPL} & 201 & \textbf{0} & 112 & \textbf{0} & 335 & \textbf{0} & 338 & \textbf{0} \\
 &                      & (17.9\%) & (\textbf{0.0\%}) & (11.8\%) & (\textbf{0.0\%}) & (35.1\%) & (\textbf{0.0\%}) & (48.5\%) & (\textbf{0.0\%}) \\
\midrule
\multirow{4}{*}{\textit{LLaMA 3B}} 
 & \multirow{2}{*}{FT}  & 133 & 96 & 113 & 101 & 85 & 76 & 60 & 35 \\
 &                      & (18.3\%) & (11.6\%) & (22.2\%) & (13.2\%) & (24.5\%) & (8.2\%) & (19.5\%) & (4.4\%) \\
\cmidrule(lr){2-10}
 & \multirow{2}{*}{CPL} & 133 & \textbf{0} & 125 & \textbf{0} & 86 & \textbf{0} & 63 & \textbf{0} \\
 &                      & (18.3\%) & (\textbf{0.0\%}) & (24.6\%) & (\textbf{0.0\%}) & (24.8\%) & (\textbf{0.0\%}) & (20.5\%) & (\textbf{0.0\%}) \\
\bottomrule
\end{tabular}
}
\end{sc}
\end{small}
\vskip -0.1in
\end{table*}

Since Conflicting Parameters are responsible for forgetting, a natural question arises: can forgetting be mitigated by freezing Conflicting Parameters and training only Collaborative Parameters? We refer to this learning paradigm as \textbf{Collaborative Parameter Learning} (CPL).

To indicate the sign of gradient similarity for the $j$-th parameter, we define indicator function $I(\cdot)$ as
\begin{equation}
I\!\left(s_{\theta_j}(\mathcal{M},\mathcal{I}) \ge 0\right)
=
\begin{cases}
1, & \text{if } s_{\theta_j}(\mathcal{M},\mathcal{I}) \ge 0 \\
0, & \text{if } s_{\theta_j}(\mathcal{M},\mathcal{I}) < 0
\end{cases}
\end{equation}
Applying the indicator function to all parameter yields
\begin{equation}
\begin{aligned}
\mathbb{I}[S(\mathcal{M},\mathcal{I}) \ge 0]
= \big(
I(s_{\theta_1}(\mathcal{M},\mathcal{I}) \ge 0), \ldots, 
I(s_{\theta_{|\theta|}}(\mathcal{M},\mathcal{I}) \ge 0)
\big)^{\top}
\end{aligned}
\end{equation}
CPL freezes parameter with negative gradient similarity, leading to a parameter update rule defined as
\begin{equation}
\theta \leftarrow \theta - \eta \, \mathbb{I}[S(\mathcal{M},\mathcal{I}) \ge 0] \odot \nabla_{\theta} L^{\mathcal{I}}(\theta)
\end{equation}
where $\odot$ denotes the Hadamard product. The change in loss on $\mathcal{M}$ after parameter update is given by
\begin{align}
\Delta L^{\mathcal{M}}(\theta) \notag
&= -\eta \, \nabla_{\theta} L^{\mathcal{M}}(\theta)^{\top}
\big(\mathbb{I}[S(\mathcal{M},\mathcal{I}) \ge 0] \odot \nabla_{\theta} L^{\mathcal{I}}(\theta)\big) \notag\\
&= -\eta \sum_{j=1}^{|\theta|}
\Big[
\nabla_{\theta_j} L^{\mathcal{M}}(\theta)\,
I(s_{\theta_j}(\mathcal{M},\mathcal{I}) \ge 0)\,
\nabla_{\theta_j} L^{\mathcal{I}}(\theta)
\Big] \notag\\
&= -\eta \sum_{j=1}^{|\theta|}
\Big[
I(s_{\theta_j}(\mathcal{M},\mathcal{I}) \ge 0)\,
s_{\theta_j}(\mathcal{M},\mathcal{I})
\Big]
\end{align}
This indicates that, given full access to the samples in $\mathcal{M}$, under a first-order approximation and with an infinitesimally small learning rate, CPL guarantees that the loss on $\mathcal{M}$ does not increase during knowledge injection.

To experimentally evaluate the effectiveness of CPL, we performed knowledge injection using both CPL and conventional fine tuning (FT) under an ideal situation where $\mathcal{M}$ can be fully accessed. The case where $\mathcal{M}$ is not explicitly available will be discussed in Section \ref{sec:out}.

CPL and FT are compared under the premise that CPL learns no fewer questions than FT. Table~\ref{tab:ft_vs_cbl} shows that FT forgets a number of mastered samples comparable to the amount of newly injected knowledge, whereas CPL forgets zero samples when $\mathcal{M}$ is fully available. This demonstrates that CPL effectively mitigates forgetting while maintaining strong capability for knowledge injection. 

\subsection{Comparison with Baseline Methods}

\begin{table*}[t]
\centering
\caption{Comparison with baseline methods on \textit{Qwen2.5 1.5B}. Results are reported in the format of ``Number of questions (Percentage)''. CPL maintains stronger capability both for knowledge injection and forgetting mitigating.}
\label{tab:baseline_comparison}
\begin{small}
\begin{sc}
\resizebox{\textwidth}{!}{
\begin{tabular}{lcccccccc}
\toprule
\multirow{2}{*} & \multicolumn{2}{c}{\textbf{MMLU}} & \multicolumn{2}{c}{\textbf{MedQA}} & \multicolumn{2}{c}{\textbf{ARC-C}} & \multicolumn{2}{c}{\textbf{CSQA}} \\
\cmidrule(lr){2-3} \cmidrule(lr){4-5} \cmidrule(lr){6-7} \cmidrule(lr){8-9}
 & \textbf{Learned} & \textbf{Forgot} & \textbf{Learned} & \textbf{Forgot} & \textbf{Learned} & \textbf{Forgot} & \textbf{Learned} & \textbf{Forgot} \\
\midrule
\multirow{2}{*}{EWC \cite{kirkpatrick2017overcoming}} & 164 & 178 & 165 & 209 & 87 & 163 & 97 & 70 \\
 & (24.2\%) & (20.3\%) & (23.4\%) & (36.8\%) & (26.9\%) & (17.2\%) & (35.7\%) & (8.5\%) \\
\cmidrule(lr){1-9}
\multirow{2}{*}{MAS \cite{aljundi2018memory}} & 167 & 200 & 184 & 259 & 104 & 80 & 86 & 161 \\
 & (24.7\%) & (22.8\%) & (26.1\%) & (45.6\%) & (32.1\%) & (8.5\%) & (31.6\%) & (19.6\%) \\
\cmidrule(lr){1-9}
\multirow{2}{*}{MOFO \cite{chen2025mofo}} & 167 & 178 & 193 & 268 & 103 & 92 & 87 & 165 \\
 & (24.7\%) & (20.3\%) & (27.4\%) & (47.2\%) & (31.8\%) & (9.7\%) & (32.0\%) & (20.0\%) \\
\cmidrule(lr){1-9}
\multirow{2}{*}{WSC \cite{cho2025forget}} & 115 & 97 & 113 & 116 & 76 & 44 & 68 & 96 \\
 & (17.0\%) & (11.0\%) & (16.1\%) & (20.4\%) & (23.5\%) & (4.7\%) & (25.0\%) & (11.7\%) \\
\cmidrule(lr){1-9}
\multirow{2}{*}{GPM \cite{saha2021gradient}} & 165 & 153 & 190 & 257 & 102 & 73 & 83 & 120 \\
 & (24.4\%) & (17.4\%) & (27.0\%) & (45.2\%) & (31.5\%) & (7.7\%) & (30.5\%) & (14.6\%) \\
\cmidrule(lr){1-9}
\multirow{2}{*}{OGD \cite{farajtabar2020orthogonal}} & 128 & 6 & 142 & 1 & 110 & 3 & 84 & 1 \\
 & (18.9\%) & (0.7\%) & (20.2\%) & (0.2\%) & (34.0\%) & (0.3\%) & (30.9\%) & (0.1\%) \\
\cmidrule(lr){1-9}
\multirow{2}{*}{A-GEM \cite{chaudhry2019efficient}} & 129 & 4 & 155 & 1 & 110 & 3 & 84 & 1 \\
 & (19.1\%) & (0.5\%) & (22.0\%) & (0.2\%) & (34.0\%) & (0.3\%) & (30.9\%) & (0.1\%) \\
\cmidrule(lr){1-9}
\multirow{2}{*}{\textbf{CPL}} & \textbf{168} & \textbf{0} & \textbf{193} & \textbf{0} & \textbf{163} & \textbf{0} & \textbf{147} & \textbf{0} \\
 & (\textbf{24.8\%}) & (\textbf{0.0\%}) & (\textbf{27.4\%}) & (\textbf{0.0\%}) & (\textbf{50.3\%}) & (\textbf{0.0\%}) & (\textbf{54.0\%}) & (\textbf{0.0\%}) \\
\bottomrule
\end{tabular}
}
\end{sc}
\end{small}
\vskip -0.1in
\end{table*}

\begin{table}[t]
\centering
\caption{Comparison of computational cost on \textit{Qwen2.5 1.5B}. CPL replaces the high-precision vector projection with low-precision element-wise sign comparison, reducing peak GPU VRAM by approximately 4.3 GB and time per epoch by 16.5\% compared to A-GEM.}
\label{tab:computational_cost}
\begin{small}
\begin{sc}
\begin{tabular}{lcc}
\toprule
& \textbf{Peak GPU VRAM (G)} & \textbf{Time / Epoch (s)} \\
\midrule
EWC & 29.5 & 3867 \\
MAS & 31.9 & 4836 \\
MOFO & 20.1 & 2170 \\
WSC & 53.0 & 1808 \\
GPM & 14.9 & 7504 \\
OGD & 20.6 & 3738 \\
A-GEM & 20.6 & 3508 \\
\textbf{CPL} & \textbf{16.3 ($\downarrow$ 4.3 GB)}  & \textbf{2928 ($\downarrow$ 16.5\%)}  \\
\bottomrule
\end{tabular}
\end{sc}
\end{small}
\vskip -0.1in
\end{table}

We further compare CPL with seven baseline methods: EWC \cite{kirkpatrick2017overcoming}, MAS \cite{aljundi2018memory}, MOFO \cite{chen2025mofo} and WSC \cite{cho2025forget}, GPM \cite{saha2021gradient}, OGD \cite{farajtabar2020orthogonal}, and A-GEM \cite{chaudhry2019efficient}. Table~\ref{tab:baseline_comparison} demonstrates that gradient-based methods (OGD and A-GEM) exhibit significantly stronger forgetting mitigation than the other baseline methods. Moreover, they are also more conceptually similar to CPL. Therefore, we focus on the comparison between CPL and gradient-based approaches. 

In terms of forgetting mitigation, CPL slightly outperforms gradient-based methods, and significantly surpasses other methods. In terms of knowledge acquisition, unlike traditional gradient-based methods, which require the gradients of $\mathcal{M}$ and $\mathcal{I}$ to be completely orthogonal, CPL masks only the conflicting component while preserving collaborative component, resulting 20.2\%-48.2\% more injected samples than OGD and A-GEM within the specified 25 epochs. 

The computational cost is compared in Table~\ref{tab:computational_cost}. Gradient-methods require high-cost vector storage for orthogonalization (FP32, accounts for 4 GB per billion parameters), whereas CPL performs element-wise sign comparison and only requires low-costs vector storage (INT8, accounts for 1 GB per billion parameters). Therefore, CPL reduces peak GPU VRAM by approximately 3 GB per billion parameters, which is confirmed by the 4.3 GB GPU VRAM reduction observed in Table~\ref{tab:computational_cost} for \textit{Qwen2.5 1.5B}. Additionally, the simplification from high-cost vector projection to element-wise sign comparison leads to a 16.5\% reduction in computational overhead (from 3508 s per epoch for OGD to 2928 s per epoch for CPL).

\subsection{Collaborative Parameters Consumption Analysis}

Since CPL freezes Conflicting Parameters and only updates Collaborative Parameters, it is important to learn that whether Collaborative Parameters are adequately preserved across different datasets. To answer this question, the change in Collaborative Parameters for each dataset is recorded after training on a specific dataset. 

Table~\ref{tab:neuron_delta} shows that Collaborative Parameters are task-dependent. After training on a dataset, the Collaborative Parameters for that dataset decrease, but the Collaborative Parameters for the other datasets even increase. This indicates that CPL does not simply consume a fixed global pool of parameters; instead, the set of Collaborative Parameters changes with the injected knowledge.

Moreover, CPL is conducted on larger-scale datasets to investigate whether Collaborative Parameters are sufficient for knowledge injection. In Appendix~\ref{app:joint_multi_dataset}, joint multi-dataset learning is conducted. Four datasets are mixed together and trained simultaneously. Table~\ref{tab:mmac_comparison} shows that FT causes 91.7\% forgetting on mixed dataset, while CPL forgets 0.0\% questions. In Appendix~\ref{app:sequential_multi_dataset}, four datasets are trained sequentially to conduct sequential multi-dataset learning. Table~\ref{tab:four_task_continual} shows that CPL nearly do not forget mastered questions across all learning orders, while FT suffers significant forgetting. 

Additionally, the learning dynamics of CPL and FT are analyzed in Appendix~\ref{app:dynamic}. Although CPL freezes Conflicting Parameters and only updates Collaborative Parameters, the speed of knowledge injection for CPL is not noticeably slowed than conventionally FT, demonstrating adequate Collaborative Parameters for training.

\section{Extending CPL to Diverse Scenarios}

\subsection{Extending CPL to More Optimizers}

In Appendix~\ref{app:theoretical}, we extend CPL to a broader family of optimizers, including Momentum, Adam, and AdamW. Experimental results in Table~\ref{tab:optimizers} prove CPL forgets zero questions across different optimizers under the ideal situation that $\mathcal{M}$ is available. Moreover, CPL is also theoretically extended to arbitrary optimizers in Appendix~\ref{app:general_optimizers}.

\subsection{Extending CPL to Various Tasks}

In Appendix~\ref{app:continual_learning}, CPL is further extended to diverse tasks including: sentiment classification (Yahoo Answers, AG News, and DBPedia) \cite{zhang2015character}, open-ended question answering (NumGLUE-cm and NumGLUE-ds) \cite{mishra2022numglue}, multilingual settings (C-STANCE) \cite{zhao2023cstance}, and multimodal tasks (MMBench) \cite{liu2024mmbench}. Table~\ref{tab:twelve_task_order1} demonstrates that CPL effectively mitigates forgetting across diverse tasks.

\subsection{Extending CPL to Out-of-Set Setting}
\label{sec:out}

\begin{table}[t]
\centering
\caption{Change in Collaborative Parameters after training. The dataset in row is the training dataset, and the dataset in column is the evaluation dataset. After training on a specific dataset, its own Collaborative Parameters decrease, but Collaborative Parameters for the others do not decrease.}
\label{tab:neuron_delta}
\begin{small}
\begin{sc}
\begin{tabular}{lcccc}
\toprule
\textbf{DATASET} & \textbf{MMLU} & \textbf{MedQA} & \textbf{ARC-C} & \textbf{CSQA} \\
\midrule
MMLU & $\textbf{-12.1\%}$ & $+20.8\%$ & $+14.3\%$ & $+13.3\%$ \\
MedQA & $+8.8\%$  & $\textbf{-8.2\%}$  & $+11.1\%$ & $+11.0\%$ \\
ARC-C & $+13.9\%$ & $+17.6\%$ & $\textbf{-10.5\%}$ & $+14.1\%$ \\
CSQA  & $+12.4\%$ & $+23.7\%$ & $+10.4\%$ & $\textbf{-9.6\%}$  \\
\bottomrule
\end{tabular}
\end{sc}
\end{small}
\vskip -0.1in
\end{table}

In the previous sections, we conducted experiments under an ideal setting that all samples in $\mathcal{M}$ are accessible. Herein, CPL is extended to out-of-set generalization settings where only a subset of $\mathcal{M}$ is accessible in Appendix~\ref{app:out_of_set_generalization}. As shown in Table~\ref{tab:held_out_eval}, when $\mathcal{M}$ is partially accessible, although CPL forgets more question than the ideal setting, it still mitigates forgetting by 59.1\%--81.7\% than conventional FT. The result also indicates that incorporating the available subset using CPL is more effective in forgetting mitigation than using replay.

\subsection{Extending CPL to Cross-Prompt Setting}

To study CPL mereley injects specific prompts or the underlying knowledge, four prompts are designed for each question as shown in Appendix~\ref{app:cross_prompt}. CPL is trained on two prompts and evaluated on the other two prompts. Table~\ref{tab:cross_prompt} shows that CPL significantly mitigates forgetting on unseen prompts.

\section{Related Work}

\subsection{Knowledge Injection}
Knowledge injection methods are commonly divided into parametric and non-parametric approaches \cite{doan2021theoretical,kotha2023understanding}. Parametric methods, such as instruction tuning, update model weights, but can overwrite previously acquired knowledge \cite{singhal2023llmmed,han2023medalpaca,qiu2024multilingual}. Non-parametric methods, such as retrieval-augmented generation, avoid parameter updates by using external evidence at inference time \cite{lewis2020rag,ram2023incontext,xiong2024medrag}, but remain limited by retrieval quality \cite{salemi2024evaluating,zhu2024information,kim2025connecting}.

\subsection{Mitigating Catastrophic Forgetting}
For model adaptation, forgetting is typically mitigated through replay, constraints, localization, or update selection \cite{de2021continual}. Replay mixes old and new data but introduces storage and privacy costs \cite{rolnick2019experience}; constraint methods such as PEFT, EWC, and MAS restrict updates or protect important weights \cite{wang2021kadapter,hu2022lora,kirkpatrick2017overcoming,aljundi2018memory}; localization and editing methods target task-relevant regions \cite{meng2022locate,meng2022mass}; and recent methods such as MOFO and WSC select or consolidate updates during adaptation \cite{chen2025mofo,cho2025forget}. Gradient similarity provides a direct signal of interference between learning objectives \cite{yu2020gradient,liu2021conflict}: conflicting gradients can degrade prior knowledge and induce forgetting \cite{farajtabar2020orthogonal}. Prior projection or alignment methods, including OGD, A-GEM, and GPM, mitigate such conflicts at a global level \cite{farajtabar2020orthogonal,chaudhry2019efficient,saha2021gradient,wang2025continual}.

\section*{Limitations}

First, our analysis relies on a first-order approximation. However, LLMs are generally non-convex. Future work could explore non-linear dynamics of forgetting. Second, CPL assumes access to $\mathcal{M}$. Although we show that CPL can still mitigate forgetting when only a subset of $\mathcal{M}$ is accessible, it remains an open question how to best estimate gradient similarity without direct access to $\mathcal{M}$. Future work could explore proxy methods for gradient estimation. Third, the gradient computation of $\mathcal{M}$ requires additional computational cost. Although CPL reduces computational cost compared with other existing gradient-based methods, it still incurs more cost than conventional FT. Future work could explore more efficient approximations of gradient similarity.

\section*{Conclusion}

Catastrophic forgetting during knowledge injection prevents LLMs from retaining previously learned knowledge. We show that strongly negative gradient similarity drives forgetting, and categorize parameters into Conflicting and Collaborative Parameters. CPL freezes Conflicting Parameters and updates only Collaborative Parameters. Experiments show that CPL significantly mitigates forgetting while maintaining knowledge injection capability across diverse scenarios.


\section*{Impact Statement}

This study provides parameter-level understanding for forgetting during knowledge injection and reveals the role of Conflicting and Collaborative parameters. Building on this insight, CPL is proposed by freezing Conflicting Parameters and updating only Collaborative Parameters. Like other knowledge injection methods, CPL may inject incorrect, biased, or harmful knowledge if the injected knowledge is poorly curated, so practical deployment requires careful data governance.

\bibliographystyle{unsrtnat}
\bibliography{references}

\newpage
\appendix
\onecolumn

\section{Dataset Details}
\label{app:details}

MMLU, MedQA, ARC-C, and CSQA are employed as foundation datasets, which are 1/10, 1/10, 1/2, and 1/10 randomly subsampled from the original datasets, respectively, resulting in 1,555, 1,273, 1,270, and 1,096 questions, respectively.

Yahoo Answers, AG News, DBPedia, NumGLUE-cm, NumGLUE-ds, C-STANCE, and MMBench are used for extensive study. All datasets are randomly subsampled to 500 questions each, except for MMBench, which contains 500 Chinese and 500 English questions.

\section{Model Inference Details}
\label{app:inference}

Model inference is conducted in a Python (3.12.2) environment, with the main packages including \texttt{transformers} (4.48.0) and \texttt{torch} (2.2.0). To ensure experimental reproducibility, the temperature is set to 0. Model inference precision is set to FP32, and experiments are conducted on a single A800 GPU.

\section{Model Inference Results}
\label{app:inference_results}

To identify Mastered Set $\mathcal{M}$ and Injection Set $\mathcal{I}$, LLMs are asked to answer the questions in the datasets. Correctly answered questions are assigned to Mastered Set $\mathcal{M}$, while incorrectly answered questions are assigned to Injection Set $\mathcal{I}$. The distribution of different sets are described in Table.~\ref{tab:dataset_distribution}.

\begin{table}[h]
\centering
\caption{Number of questions in Mastered Set $\mathcal{M}$ and Injection Set $\mathcal{I}$. ``INJECTION'' denotes the number of samples in Injection Set (answered incorrectly), while ``MASTERED'' denotes the number of samples in Mastered Set (answered correctly).}
\label{tab:dataset_distribution}
\begin{small}
\begin{sc}
\resizebox{\textwidth}{!}{
\begin{tabular}{lcccccccc}
\toprule
\multirow{2}{*} & \multicolumn{2}{c}{\textbf{MMLU}} & \multicolumn{2}{c}{\textbf{MedQA}} & \multicolumn{2}{c}{\textbf{ARC-C}} & \multicolumn{2}{c}{\textbf{CSQA}} \\
\cmidrule(lr){2-3} \cmidrule(lr){4-5} \cmidrule(lr){6-7} \cmidrule(lr){8-9}
 & \textbf{Injection} & \textbf{Mastered} & \textbf{Injection} & \textbf{Mastered} & \textbf{Injection} & \textbf{Mastered} & \textbf{Injection} & \textbf{Mastered} \\
\midrule
\textit{Qwen 1.5B} & 678  & 877  & 704 & 568 & 324 & 946  & 272 & 824 \\
\textit{Qwen 3B}   & 585  & 970  & 639 & 633 & 212 & 1058 & 243 & 853 \\
\textit{Qwen 7B}   & 454  & 1101 & 502 & 770 & 134 & 1136 & 197 & 899 \\
\textit{LLaMA 1B}  & 1126 & 429  & 952 & 320 & 954 & 316  & 697 & 399 \\
\textit{LLaMA 3B}  & 727  & 828  & 508 & 764 & 347 & 923  & 308 & 788 \\
\bottomrule
\end{tabular}
}
\end{sc}
\end{small}
\vskip -0.1in
\end{table}

\section{Model Training Details}
\label{app:training}

The environment of model training is essentially the same as that for model inference. The number of epochs is set to 25. For the SGD optimizer, the learning rate included three values according to their magnitude of gradient: $2\text{e-}9$, $2\text{e-}7$, and $1\text{e-}7$ (see the code for exact configurations). For Momentum optimizer, the learning rate is consistent with SGD, and the momentum coefficient is set to 0.9. For Adam optimizer (without regularization), $\beta_1$ and $\beta_2$ are 0.9 and 0.999 respectively, and $\epsilon$ is $1\text{e-}8$. For AdamW optimizer, the regularization parameter (weight decay) is $1\text{e-}1$. Model training is performed on a single NVIDIA A800 GPU.

\section{Model Training Results}

The samples in Injection Set $\mathcal{I}$ are trained for knowledge injection. After training, the number of correctly answered questions in Injection Set $\mathcal{I}$ is employed to evaluate learning, while the number of incorrectly answered questions in Mastered Set $\mathcal{M}$ is employed to evaluate forgetting. The results in Table.~\ref{tab:learning_forgetting_details} indicates that serious catastrophic forgetting happens during knowledge injection. 

\label{app:training_results}
\begin{table}[h]
\centering
\caption{Learning and forgetting during knowledge injection. Results are presented in the format of ``Number of Items (Percentage)''. After training, although new knowledge is successfully injected into LLMs, severe catastrophic forgetting is also observed.}
\label{tab:learning_forgetting_details}
\begin{small}
\begin{sc}
\resizebox{\textwidth}{!}{
\begin{tabular}{lcccccccc}
\toprule
\multirow{2}{*} & \multicolumn{2}{c}{\textbf{MMLU}} & \multicolumn{2}{c}{\textbf{MedQA}} & \multicolumn{2}{c}{\textbf{ARC-C}} & \multicolumn{2}{c}{\textbf{CSQA}} \\
\cmidrule(lr){2-3} \cmidrule(lr){4-5} \cmidrule(lr){6-7} \cmidrule(lr){8-9}
 & \textbf{Learned} & \textbf{Forgot} & \textbf{Learned} & \textbf{Forgot} & \textbf{Learned} & \textbf{Forgot} & \textbf{Learned} & \textbf{Forgot} \\
\midrule
\multirow{2}{*}{\textit{Qwen 1.5B}} 
 & 168 & 188 & 191 & 264 & 104 & 88 & 86 & 163 \\
 & (24.8\%) & (21.4\%) & (27.1\%) & (46.5\%) & (32.1\%) & (9.3\%) & (31.6\%) & (19.8\%) \\
\midrule
\multirow{2}{*}{\textit{Qwen 3B}} 
 & 114 & 126 & 119 & 160 & 71 & 286 & 77 & 105 \\
 & (19.5\%) & (13.0\%) & (18.6\%) & (25.3\%) & (33.5\%) & (27.0\%) & (31.7\%) & (12.3\%) \\
\midrule
\multirow{2}{*}{\textit{Qwen 7B}} 
 & 127 & 193 & 110 & 169 & 62 & 19 & 68 & 48 \\
 & (28.0\%) & (17.5\%) & (21.9\%) & (21.9\%) & (46.3\%) & (1.7\%) & (34.5\%) & (5.3\%) \\
\midrule
\multirow{2}{*}{\textit{LLaMA 1B}} 
 & 158 & 73 & 105 & 43 & 328 & 154 & 333 & 176 \\
 & (14.0\%) & (17.0\%) & (11.0\%) & (13.4\%) & (34.4\%) & (48.7\%) & (47.8\%) & (44.1\%) \\
\midrule
\multirow{2}{*}{\textit{LLaMA 3B}} 
 & 133 & 96 & 113 & 101 & 85 & 76 & 60 & 35 \\
 & (18.3\%) & (11.6\%) & (22.2\%) & (13.2\%) & (24.5\%) & (8.2\%) & (19.5\%) & (4.4\%) \\
\bottomrule
\end{tabular}
}
\end{sc}
\end{small}
\vskip -0.1in
\end{table}

\section{Evaluation on Parameter-Efficient Fine-Tuning}
\label{app:lora}

We evaluate LoRA as a representative parameter-efficient fine-tuning baseline. All experiments are conducted on the \textit{Qwen2.5-1.5B}. For a fair comparison with FT, we adopt the same training protocol with Appendix~\ref{app:training}, using the SGD optimizer with a learning rate of $\eta=1\times10^{-5}$. The LoRA configuration employs a rank of $r=16$ and a scaling factor of $\alpha=32$, with LoRA modules applied to all linear layers in the attention mechanism. The batch size and number of training epochs are identical to those used in Appendix~\ref{app:training}.

As shown in Table~\ref{tab:lora_comparison}, when LoRA achieves learning performance comparable to FT, LoRA does not alleviate catastrophic forgetting, exhibiting similarly severe forgetting across all evaluated datasets.

\begin{table}[h]
\centering
\caption{Comparison of FT and LoRA during knowledge injection. Forgetting cannot be effectively mitigated when LoRA is employed.}
\label{tab:lora_comparison}
\begin{small}
\begin{sc}
\resizebox{\textwidth}{!}{
\begin{tabular}{lcccccccc}
\toprule
\multirow{2}{*} & \multicolumn{2}{c}{\textbf{MMLU}} & \multicolumn{2}{c}{\textbf{MedQA}} & \multicolumn{2}{c}{\textbf{ARC-C}} & \multicolumn{2}{c}{\textbf{CSQA}} \\
\cmidrule(lr){2-3} \cmidrule(lr){4-5} \cmidrule(lr){6-7} \cmidrule(lr){8-9}
 & \textbf{Learned} & \textbf{Forgot} & \textbf{Learned} & \textbf{Forgot} & \textbf{Learned} & \textbf{Forgot} & \textbf{Learned} & \textbf{Forgot} \\
\midrule
\multirow{2}{*}{FT}   & 168 & 188 & 191 & 264 & 104 & 88 & 86 & 163 \\
                      & (24.8\%) & (21.4\%) & (27.1\%) & (46.5\%) & (32.1\%) & (9.3\%) & (31.6\%) & (19.8\%) \\
\cmidrule(lr){1-9}
\multirow{2}{*}{LoRA} & 168 & 176 & 193 & 286 & 104 & 79 & 86 & 164 \\
                      & (24.8\%) & (20.1\%) & (27.4\%) & (50.4\%) & (32.1\%) & (8.4\%) & (31.6\%) & (19.9\%) \\
\bottomrule
\end{tabular}
}
\end{sc}
\end{small}
\vskip -0.1in
\end{table}


\section{Extending CPL to Larger-Scale Datasets}
\label{app:complex_dataseet}

\subsection{Extending CPL to Joint Multi-Dataset Learning}
\label{app:joint_multi_dataset}

Four datasets are combined to construct a more complex dataset, termed MMAC, to evaluate the performance of CPL on larger-scale datasets. Table~\ref{tab:mmac_comparison} indicates that FT induced 91.7\% forgetting during knowledge injection, while CPL forgets zero questions under the ideal situation that Mastered Set $\mathcal{M}$ is fully accessed. 

\begin{table}[h]
\centering
\caption{Performance of joint multi-dataset learning. On the larger-scale MMAC dataset, FT causes severe forgetting (91.7\%), whereas CPL maintains 0.0\% forgetting.}
\label{tab:mmac_comparison}
\begin{small}
\begin{sc}
\resizebox{\textwidth}{!}{
\begin{tabular}{lcccccccccc}
\toprule
\multirow{2}{*} & \multicolumn{2}{c}{\textbf{MMAC}} & \multicolumn{2}{c}{\textbf{MMLU}} & \multicolumn{2}{c}{\textbf{MedQA}} & \multicolumn{2}{c}{\textbf{ARC-C}} & \multicolumn{2}{c}{\textbf{CSQA}} \\
\cmidrule(lr){2-3} \cmidrule(lr){4-5} \cmidrule(lr){6-7} \cmidrule(lr){8-9} \cmidrule(lr){10-11}
 & \textbf{Learned} & \textbf{Forgot} & \textbf{Learned} & \textbf{Forgot} & \textbf{Learned} & \textbf{Forgot} & \textbf{Learned} & \textbf{Forgot} & \textbf{Learned} & \textbf{Forgot} \\
\midrule
\multirow{2}{*}{FT} 
 & 1878 & 2949 & 625 & 806 & 665 & 547 & 316 & 819 & 272 & 777 \\
 & (94.9\%) & (91.7\%) & (92.2\%) & (91.9\%) & (94.5\%) & (96.3\%) & (97.5\%) & (86.6\%) & (100\%) & (94.3\%) \\
\midrule
\multirow{2}{*}{CPL} 
 & 1893 & \textbf{0} & 604 & \textbf{0} & 700 & \textbf{0} & 321 & \textbf{0} & 268 & \textbf{0} \\
 & (95.7\%) & (\textbf{0.0\%}) & (89.1\%) & (\textbf{0.0\%}) & (99.4\%) & (\textbf{0.0\%}) & (99.1\%) & (\textbf{0.0\%}) & (98.5\%) & (\textbf{0.0\%}) \\
\bottomrule
\end{tabular}
}
\end{sc}
\end{small}
\vskip -0.1in
\end{table}

\subsection{Extending CPL to Sequential Multi-Dataset Learning}
\label{app:sequential_multi_dataset}

The datasets are trained sequentially according to the ``Order'' column in 
Table~\ref{tab:four_task_continual}. CPL consistently mitigates catastrophic forgetting across different task orders.

\begin{table*}[h]
\centering
\caption{Performance of sequential multi-dataset learning. AR, CS, MM, and Me denote ARC-C, CSQA, MMLU, and MedQA, respectively. Results indicate that CPL nearly do not forget questions across all orders while maintaining effective learning, under the ideal situtation that Mastered Set $\mathcal{M}$ is fully accessed.}
\label{tab:four_task_continual}
\begin{small}
\begin{sc}
\resizebox{\textwidth}{!}{
\begin{tabular}{lcccccccc}
\toprule
\multirow{2}{*}{\textbf{Order}} & \multicolumn{2}{c}{\textbf{MMLU}} & \multicolumn{2}{c}{\textbf{MedQA}} & \multicolumn{2}{c}{\textbf{ARC-C}} & \multicolumn{2}{c}{\textbf{CSQA}} \\
\cmidrule(lr){2-3} \cmidrule(lr){4-5} \cmidrule(lr){6-7} \cmidrule(lr){8-9}
 & \textbf{Learned} & \textbf{Forgot} & \textbf{Learned} & \textbf{Forgot} & \textbf{Learned} & \textbf{Forgot} & \textbf{Learned} & \textbf{Forgot} \\
\midrule
AR-CS-MM-Me & 151 & 0 & 167 & 0 & 109 & 0 & 90 & 0 \\
MM-Me-AR-CS & 141 & 1 & 149 & 0 & 113 & 0 & 94 & 0 \\
CS-AR-Me-MM & 150 & 0 & 148 & 0 & 106 & 0 & 98 & 0 \\
Me-AR-CS-MM & 145 & 0 & 152 & 0 & 110 & 0 & 92 & 0 \\
\bottomrule
\end{tabular}
}
\end{sc}
\end{small}
\vskip -0.1in
\end{table*}


\section{Dynamic Characteristic of CPL}
\label{app:dynamic}

\begin{figure*}[t]
\centering

\begin{minipage}{0.03\textwidth} ~ \end{minipage} 
\hfill
\begin{minipage}{0.02\textwidth} ~ \end{minipage} 
\hfill
\begin{subfigure}{0.22\textwidth}
    \centering
    \textbf{\small MMLU}
\end{subfigure}
\hfill
\begin{subfigure}{0.22\textwidth}
    \centering
    \textbf{\small MedQA}
\end{subfigure}
\hfill
\begin{subfigure}{0.22\textwidth}
    \centering
    \textbf{\small ARC-C}
\end{subfigure}
\hfill
\begin{subfigure}{0.22\textwidth}
    \centering
    \textbf{\small CSQA}
\end{subfigure}

\begin{minipage}[c]{0.03\textwidth}
    \centering
    \rotatebox{90}{\textbf{\small SGD}}
\end{minipage}
\hfill
\begin{minipage}[c]{0.02\textwidth}
    \centering
    \rotatebox{90}{\textbf{\scriptsize Learning}}
    \par \vspace{3em} \par 
    \rotatebox{90}{\textbf{\scriptsize Forgetting}}
\end{minipage}
\hfill
\begin{subfigure}[c]{0.22\textwidth}
    \includegraphics[width=\linewidth]{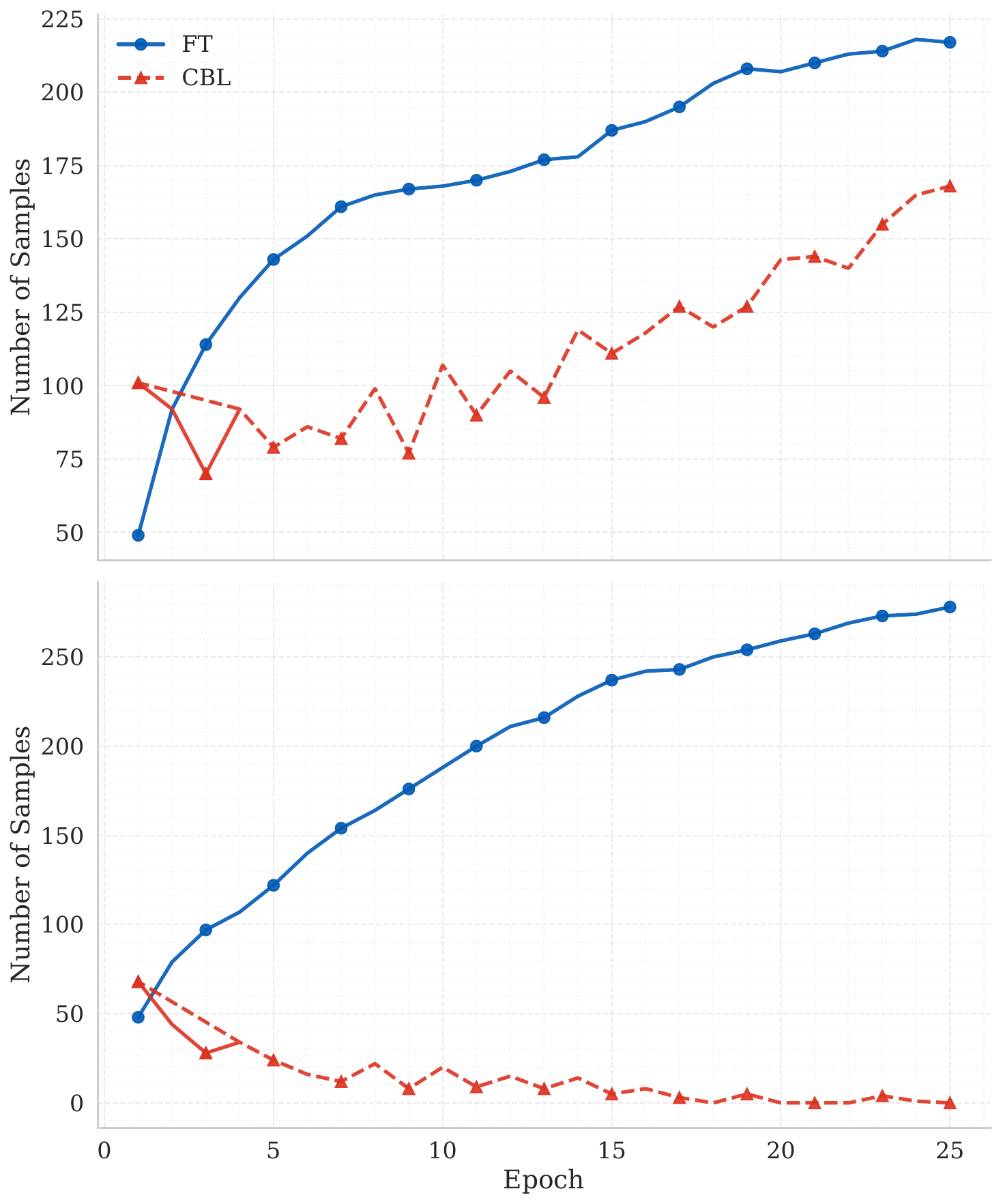}
\end{subfigure}
\hfill
\begin{subfigure}[c]{0.22\textwidth}
    \includegraphics[width=\linewidth]{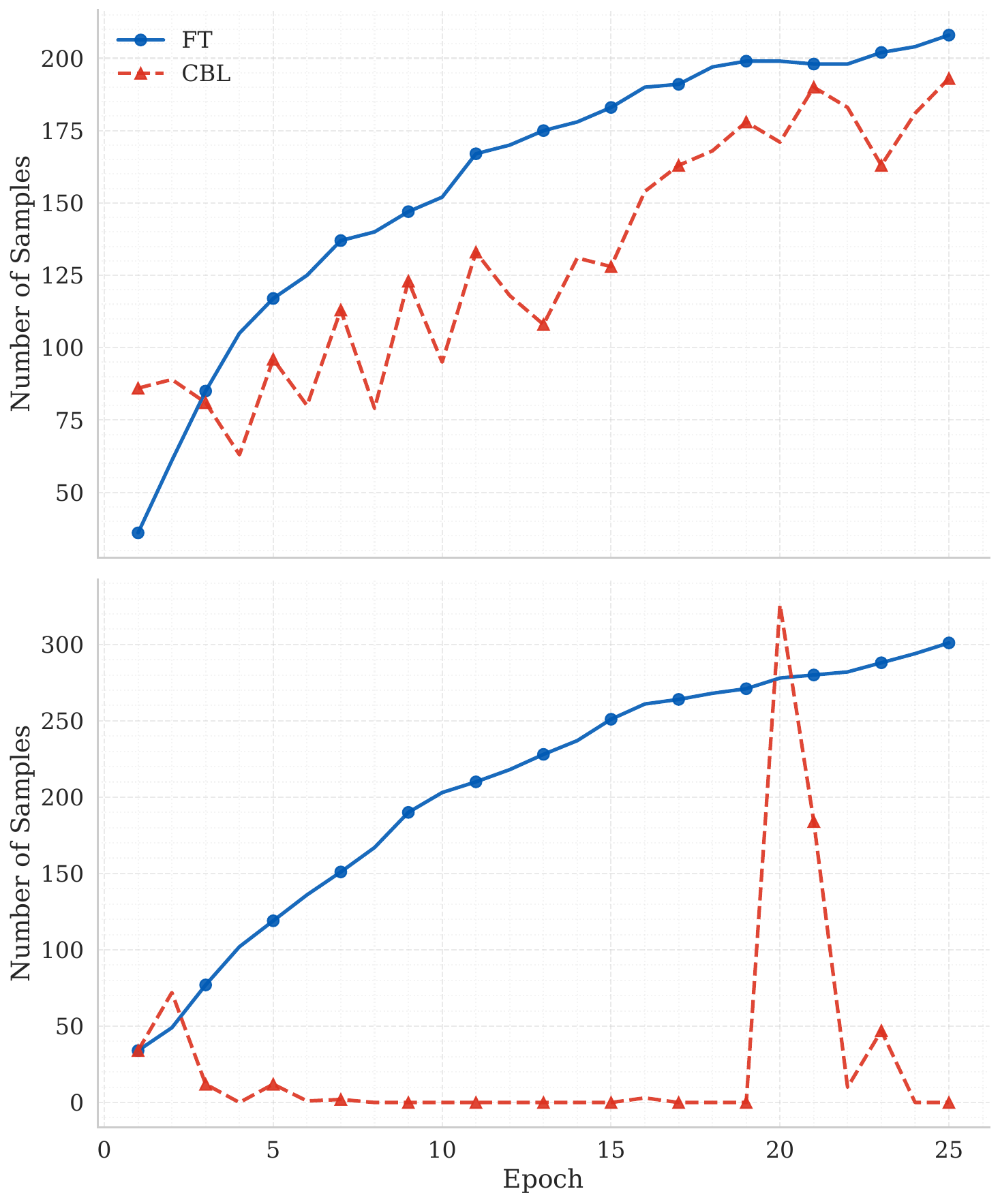}
\end{subfigure}
\hfill
\begin{subfigure}[c]{0.22\textwidth}
    \includegraphics[width=\linewidth]{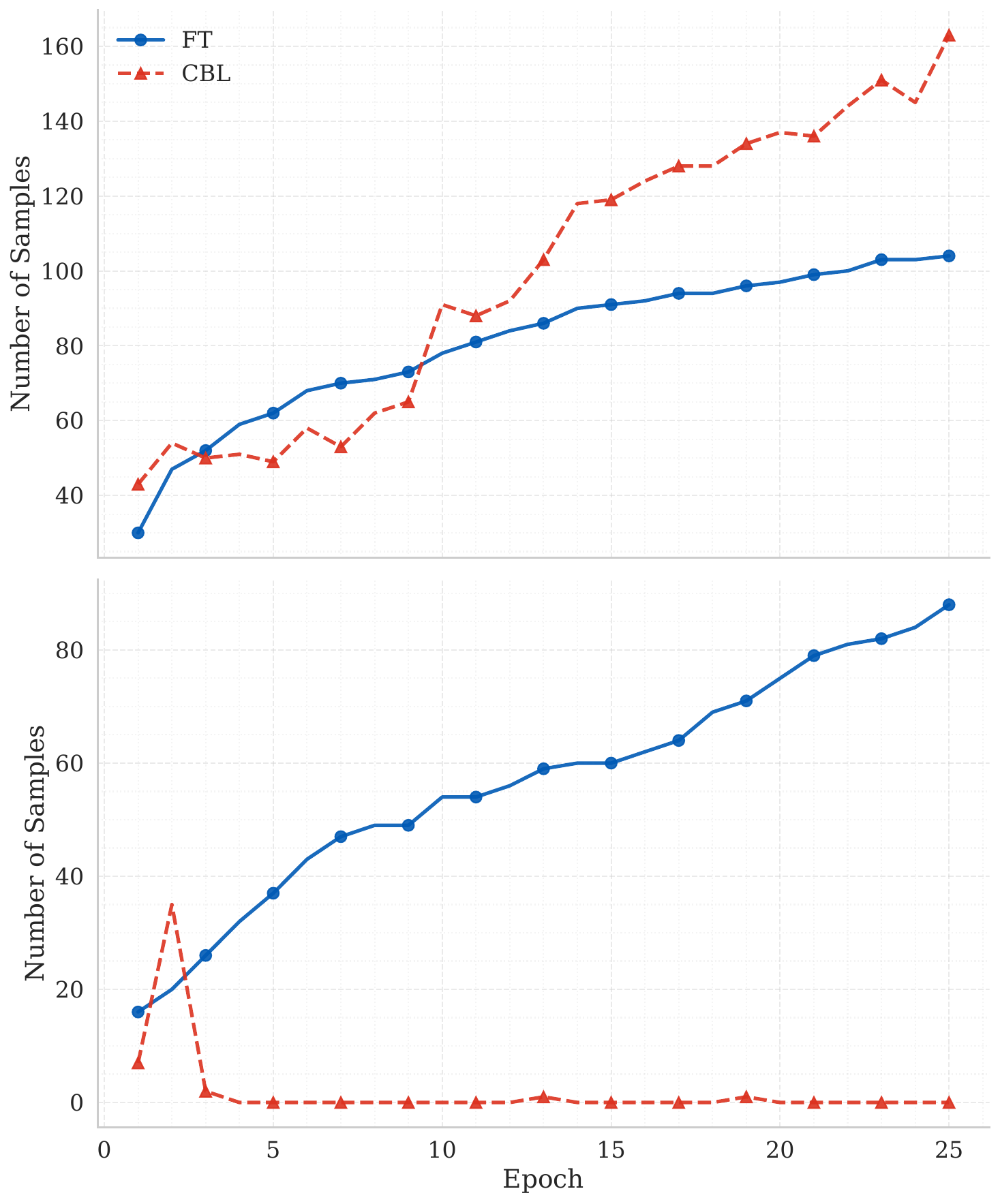}
\end{subfigure}
\hfill
\begin{subfigure}[c]{0.22\textwidth}
    \includegraphics[width=\linewidth]{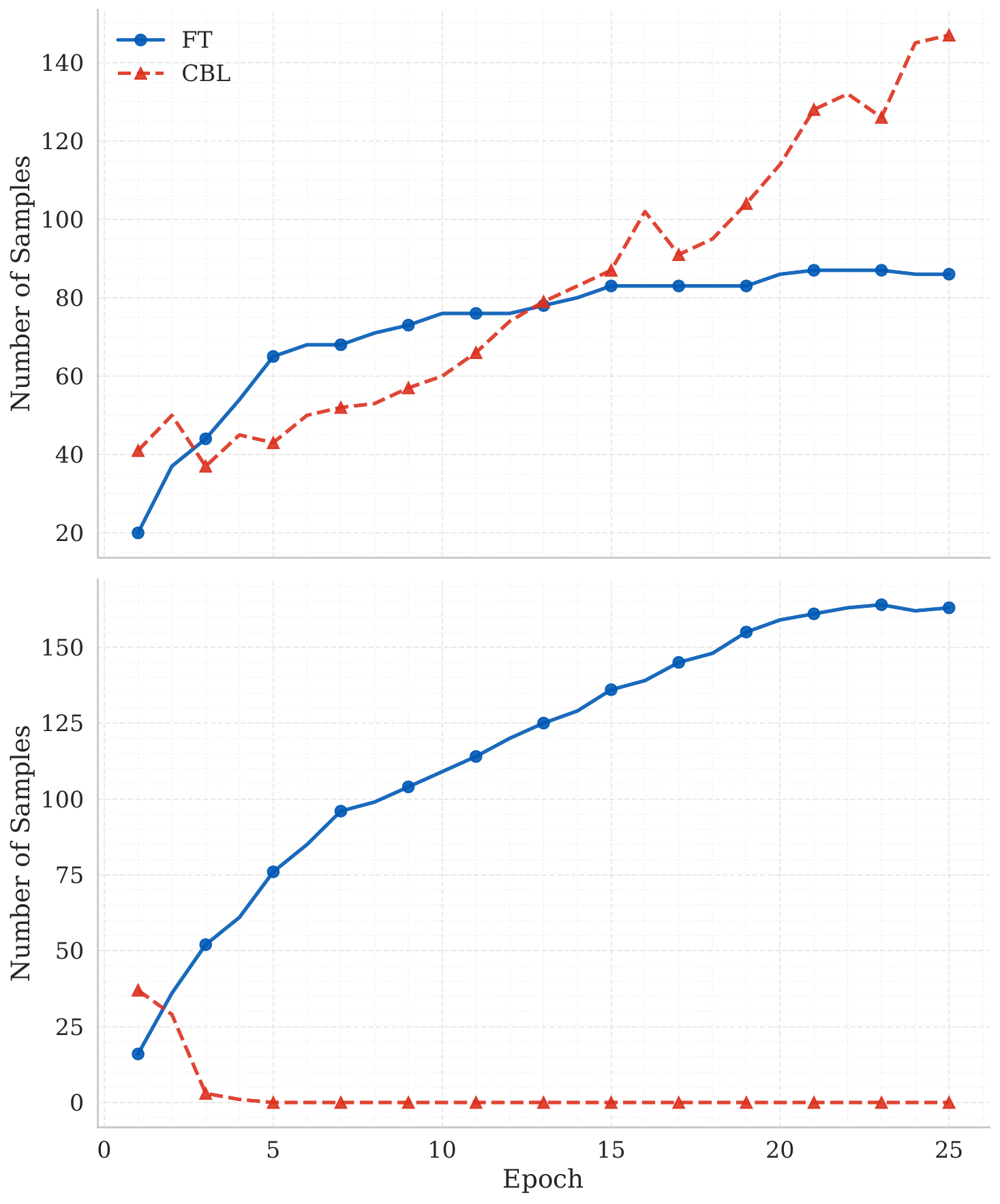}
\end{subfigure}

\begin{minipage}[c]{0.03\textwidth}
    \centering
    \rotatebox{90}{\textbf{\small Momentum}}
\end{minipage}
\hfill
\begin{minipage}[c]{0.02\textwidth}
    \centering
    \rotatebox{90}{\textbf{\scriptsize Learning}}
    \par \vspace{3em} \par 
    \rotatebox{90}{\textbf{\scriptsize Forgetting}}
\end{minipage}
\hfill
\begin{subfigure}[c]{0.22\textwidth}
    \includegraphics[width=\linewidth]{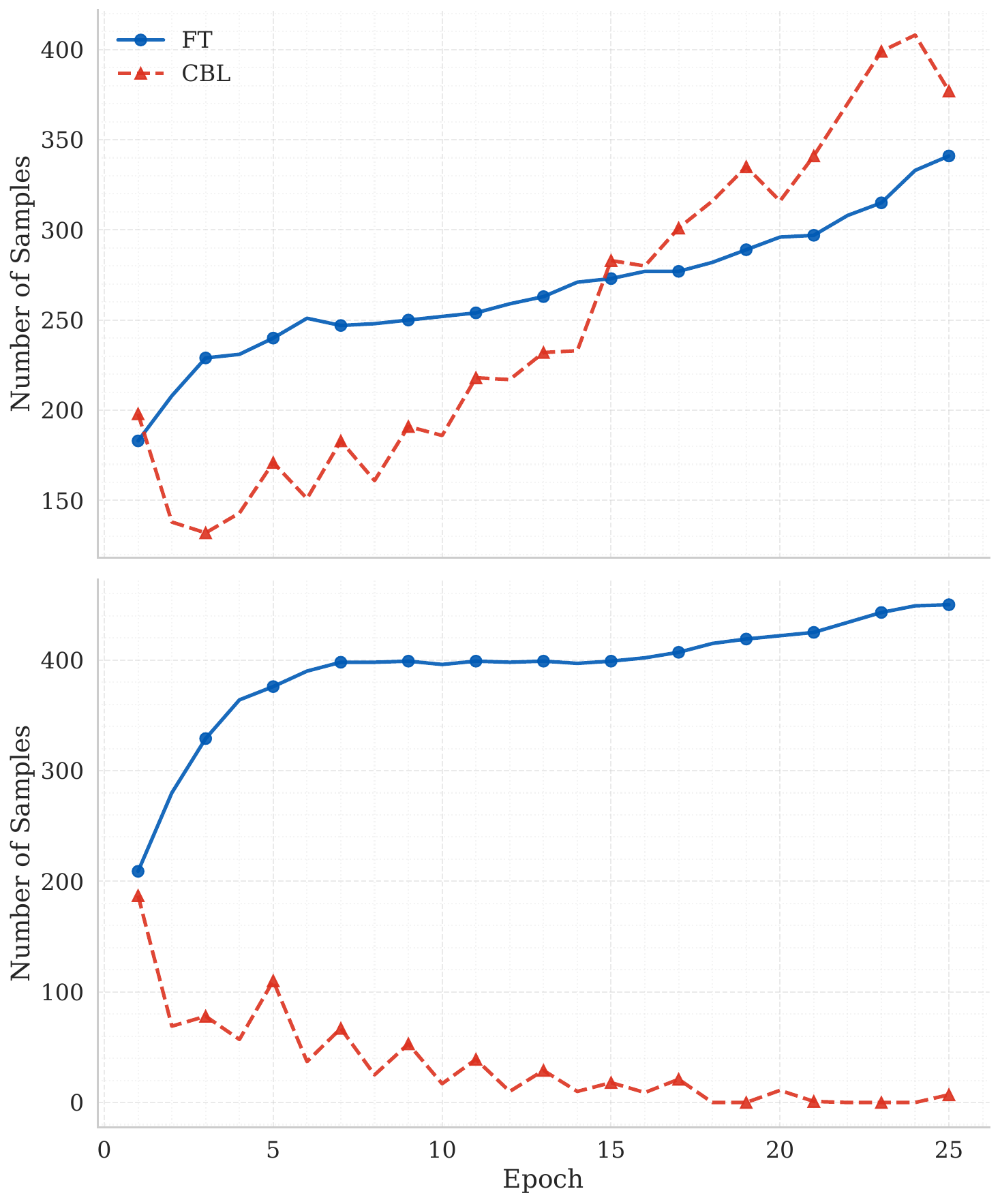}
\end{subfigure}
\hfill
\begin{subfigure}[c]{0.22\textwidth}
    \includegraphics[width=\linewidth]{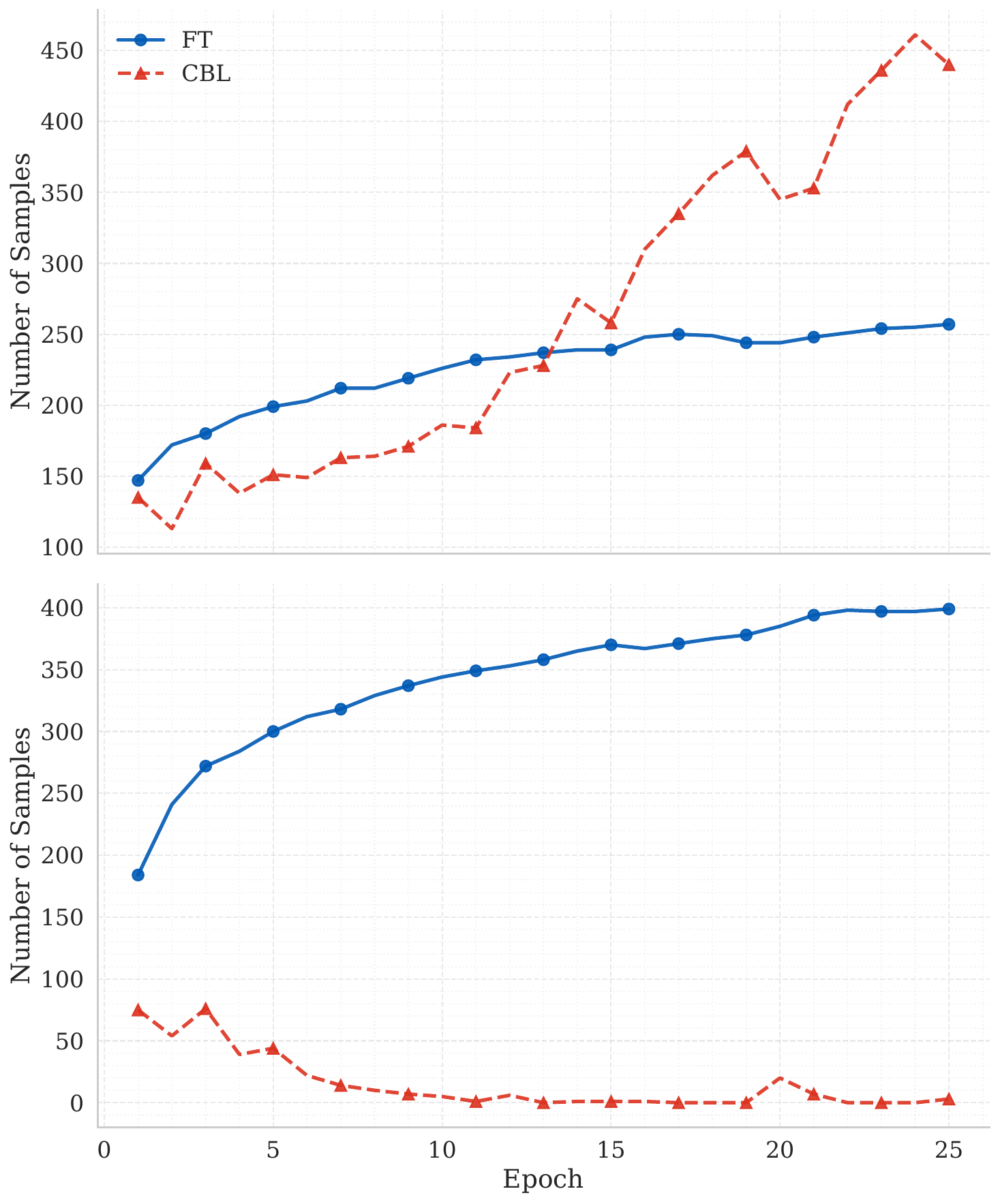}
\end{subfigure}
\hfill
\begin{subfigure}[c]{0.22\textwidth}
    \includegraphics[width=\linewidth]{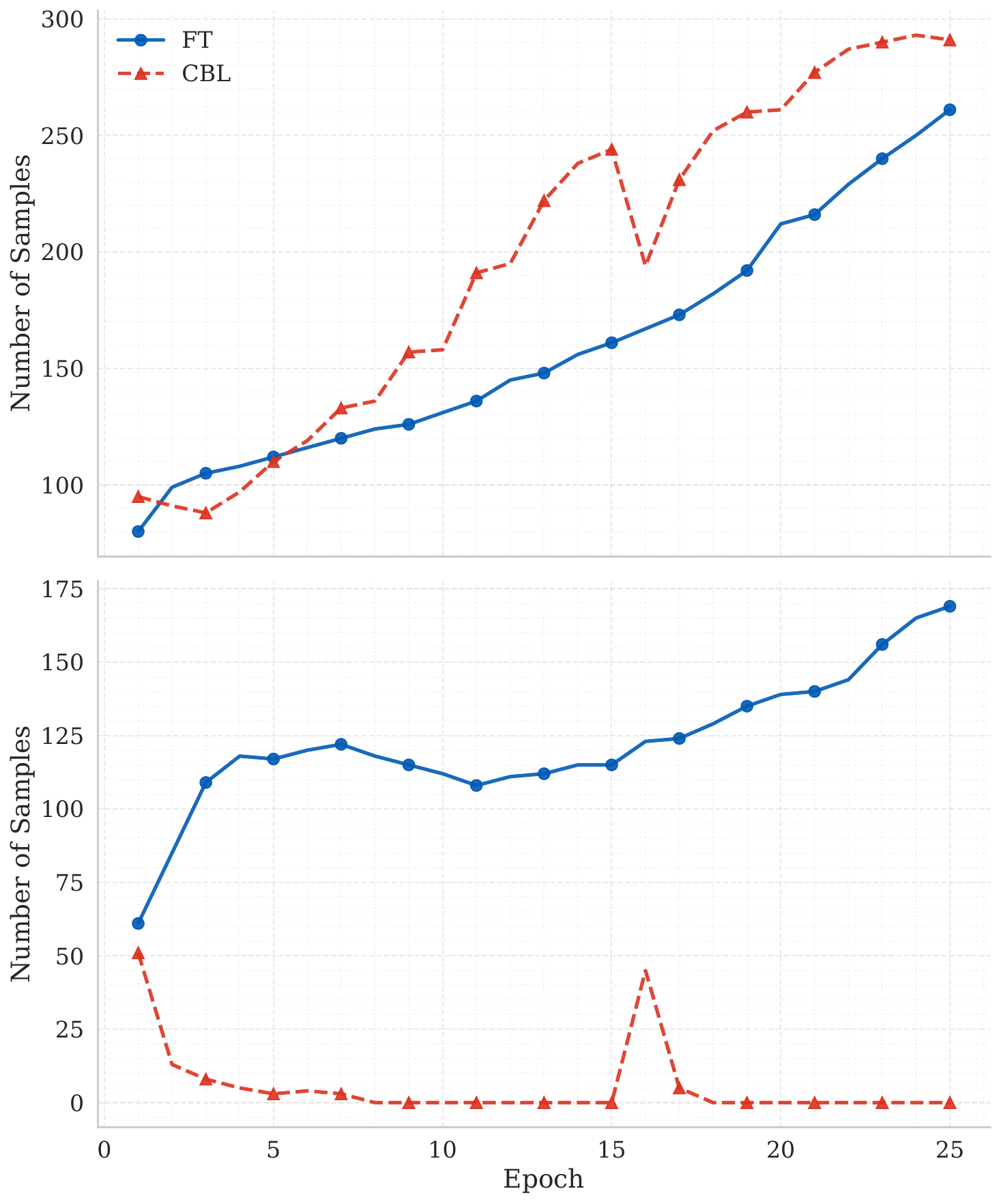}
\end{subfigure}
\hfill
\begin{subfigure}[c]{0.22\textwidth}
    \includegraphics[width=\linewidth]{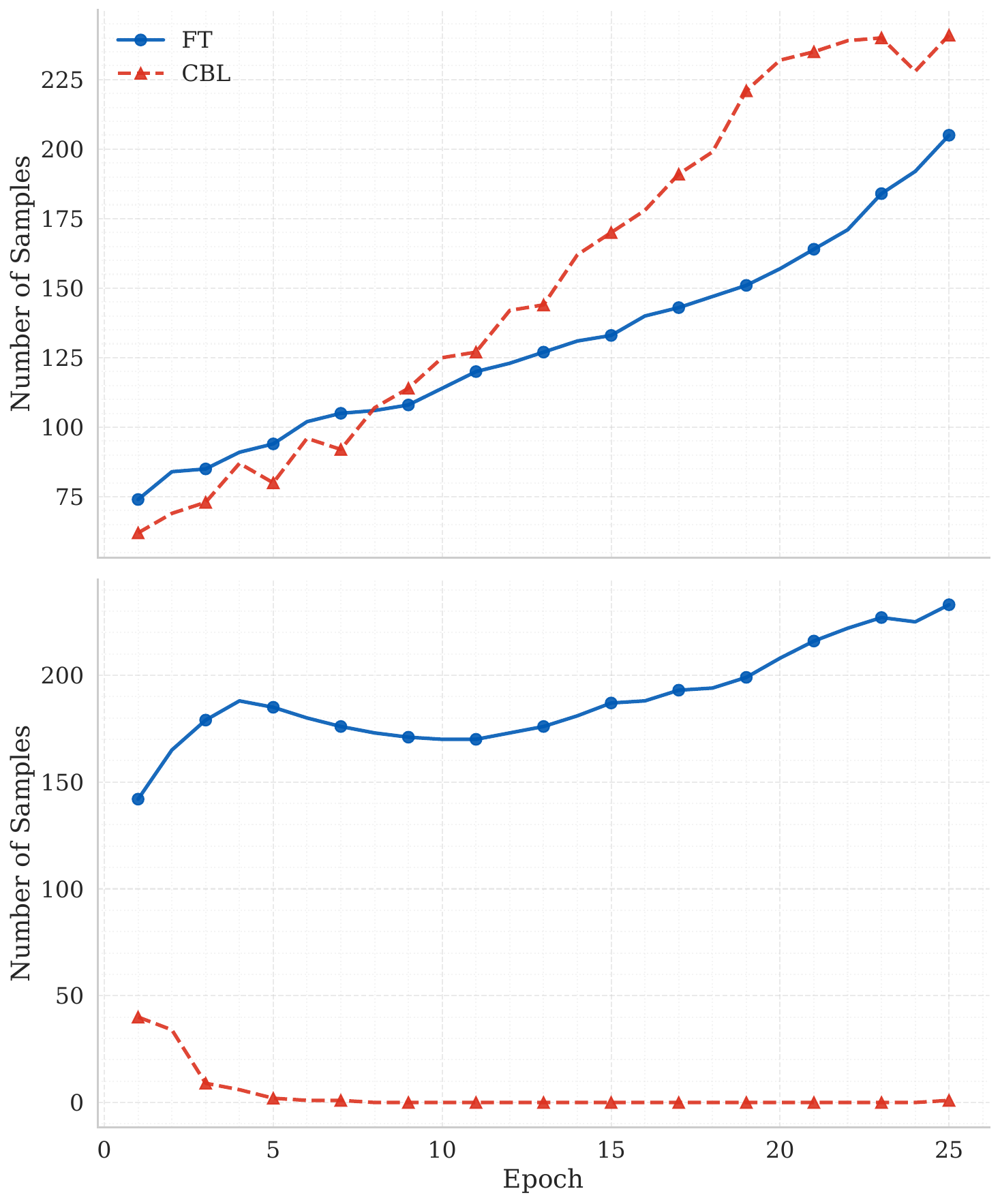}
\end{subfigure}

\begin{minipage}[c]{0.03\textwidth}
    \centering
    \rotatebox{90}{\textbf{\small Adam}}
\end{minipage}
\hfill
\begin{minipage}[c]{0.02\textwidth}
    \centering
    \rotatebox{90}{\textbf{\scriptsize Learning}}
    \par \vspace{3em} \par 
    \rotatebox{90}{\textbf{\scriptsize Forgetting}}
\end{minipage}
\hfill
\begin{subfigure}[c]{0.22\textwidth}
    \includegraphics[width=\linewidth]{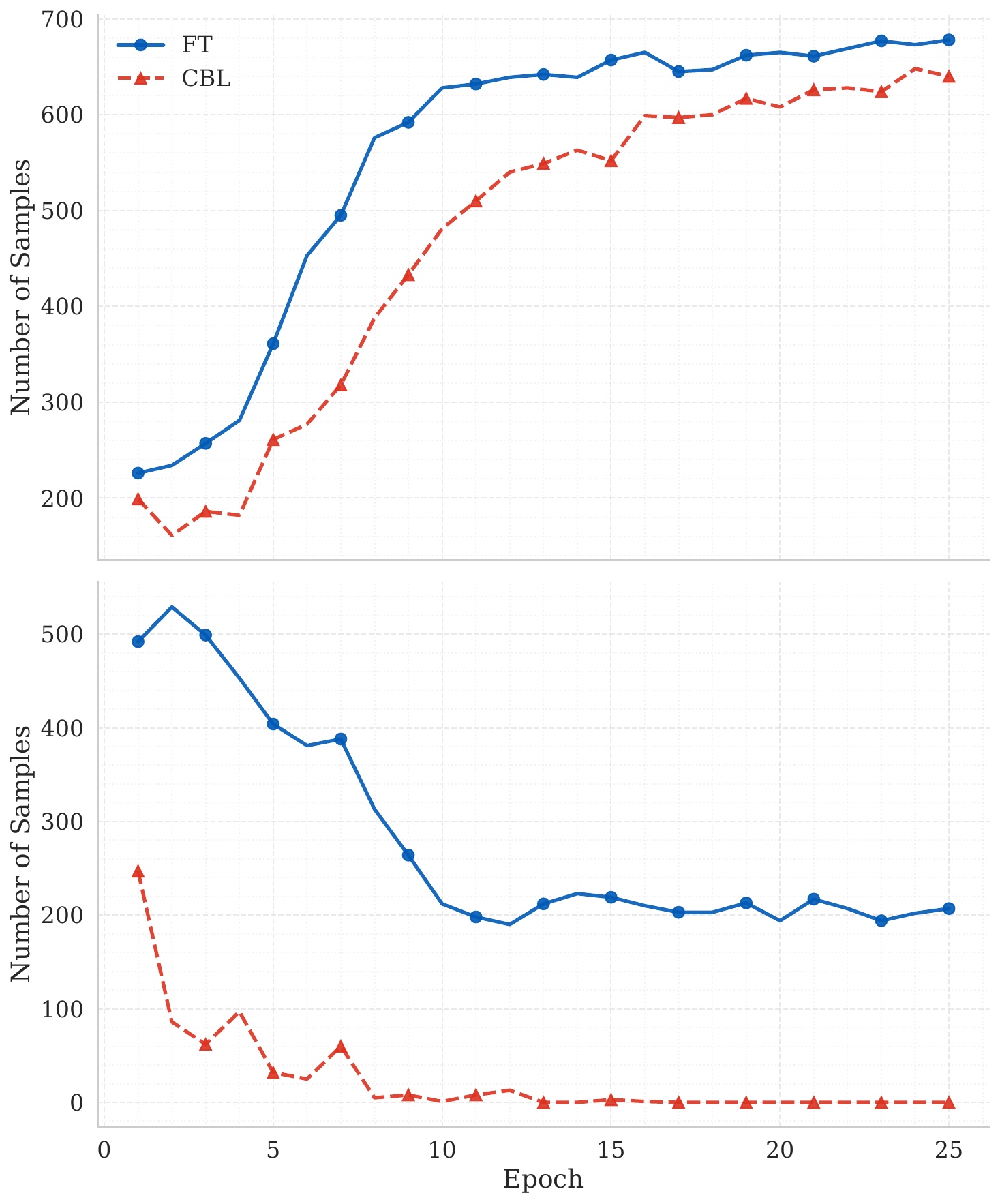}
\end{subfigure}
\hfill
\begin{subfigure}[c]{0.22\textwidth}
    \includegraphics[width=\linewidth]{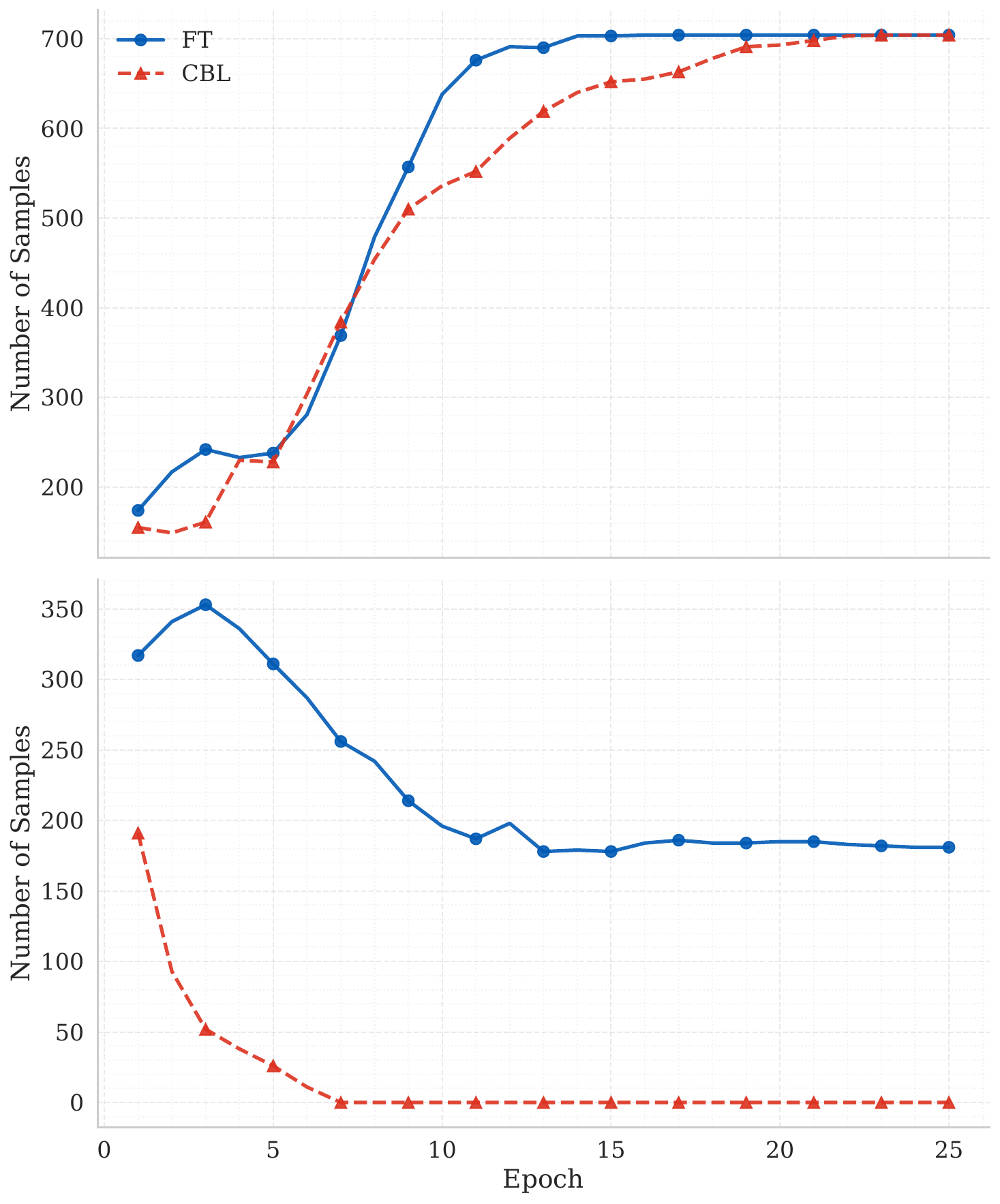}
\end{subfigure}
\hfill
\begin{subfigure}[c]{0.22\textwidth}
    \includegraphics[width=\linewidth]{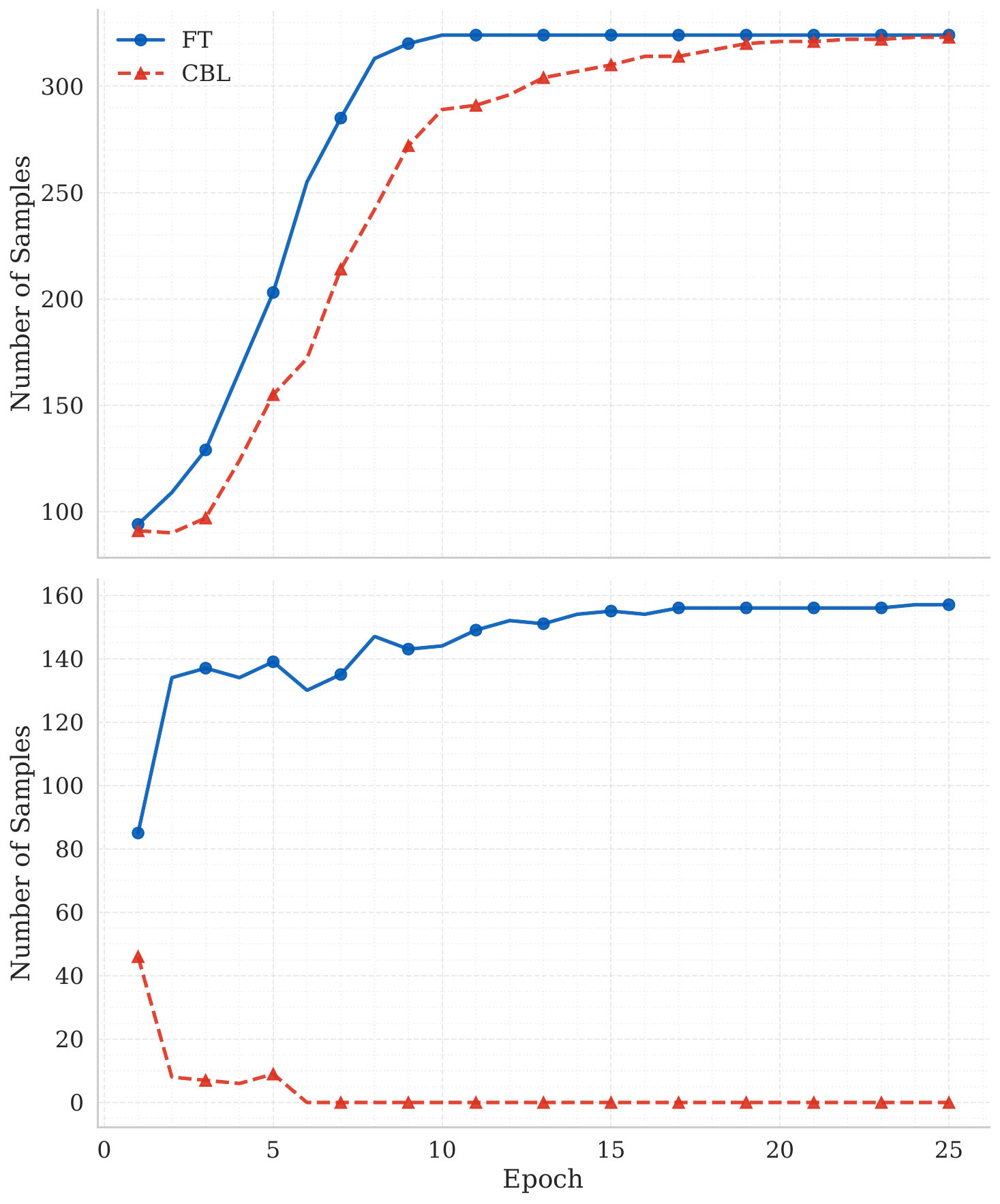}
\end{subfigure}
\hfill
\begin{subfigure}[c]{0.22\textwidth}
    \includegraphics[width=\linewidth]{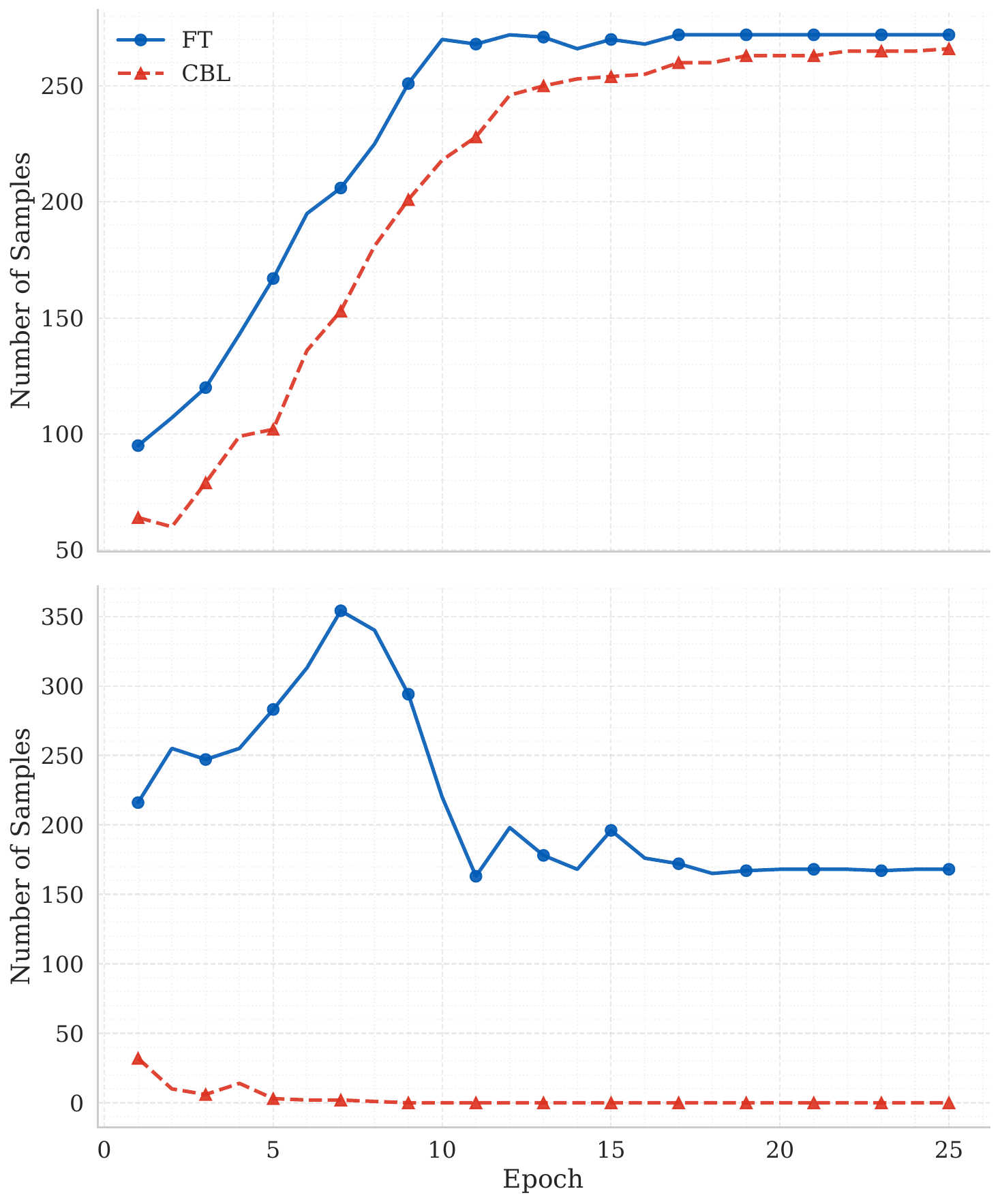}
\end{subfigure}

\begin{minipage}[c]{0.03\textwidth}
    \centering
    \rotatebox{90}{\textbf{\small AdamW}}
\end{minipage}
\hfill
\begin{minipage}[c]{0.02\textwidth}
    \centering
    \rotatebox{90}{\textbf{\scriptsize Learning}}
    \par \vspace{3em} \par 
    \rotatebox{90}{\textbf{\scriptsize Forgetting}}
\end{minipage}
\hfill
\begin{subfigure}[c]{0.22\textwidth}
    \includegraphics[width=\linewidth]{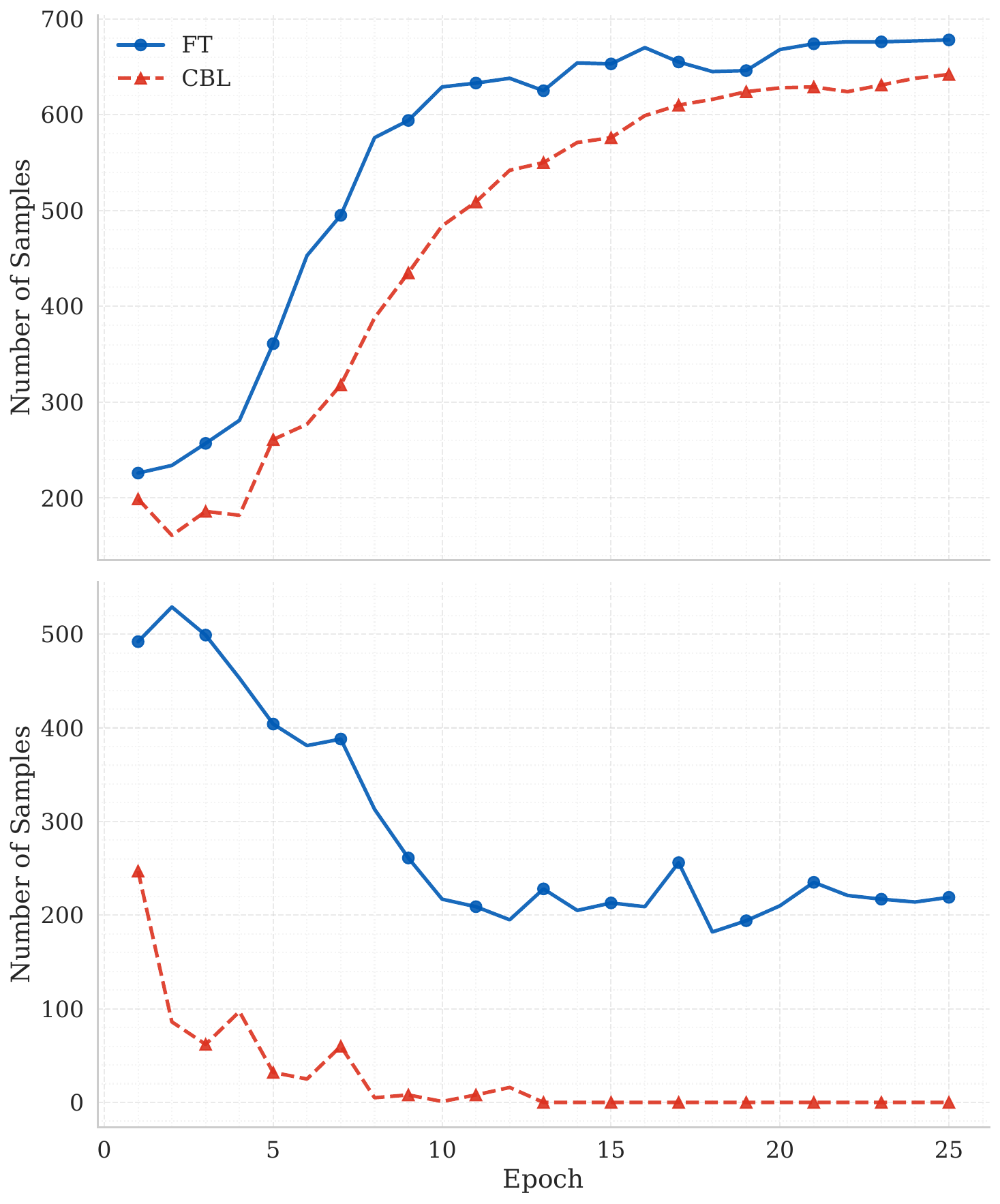}
\end{subfigure}
\hfill
\begin{subfigure}[c]{0.22\textwidth}
    \includegraphics[width=\linewidth]{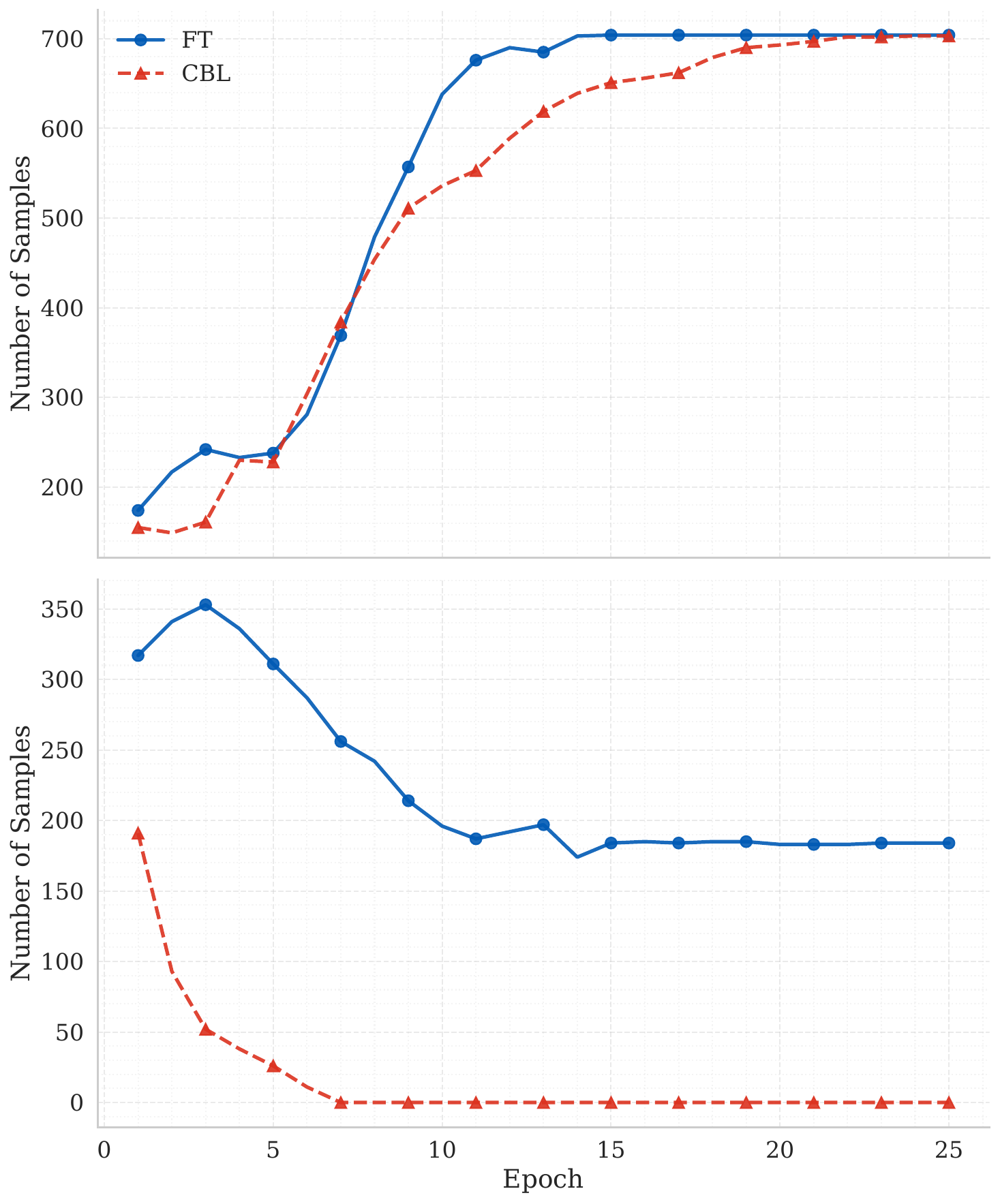}
\end{subfigure}
\hfill
\begin{subfigure}[c]{0.22\textwidth}
    \includegraphics[width=\linewidth]{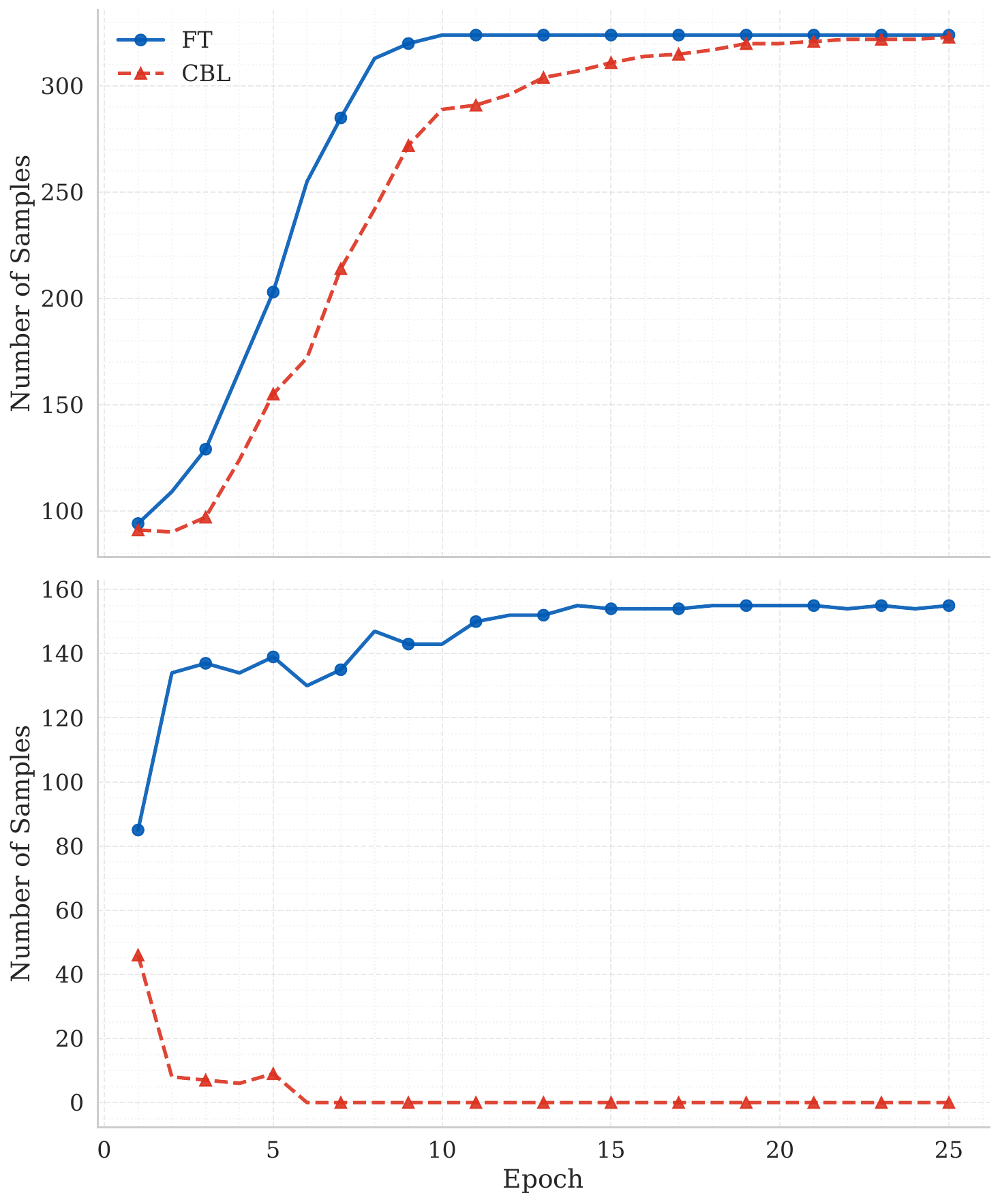}
\end{subfigure}
\hfill
\begin{subfigure}[c]{0.22\textwidth}
    \includegraphics[width=\linewidth]{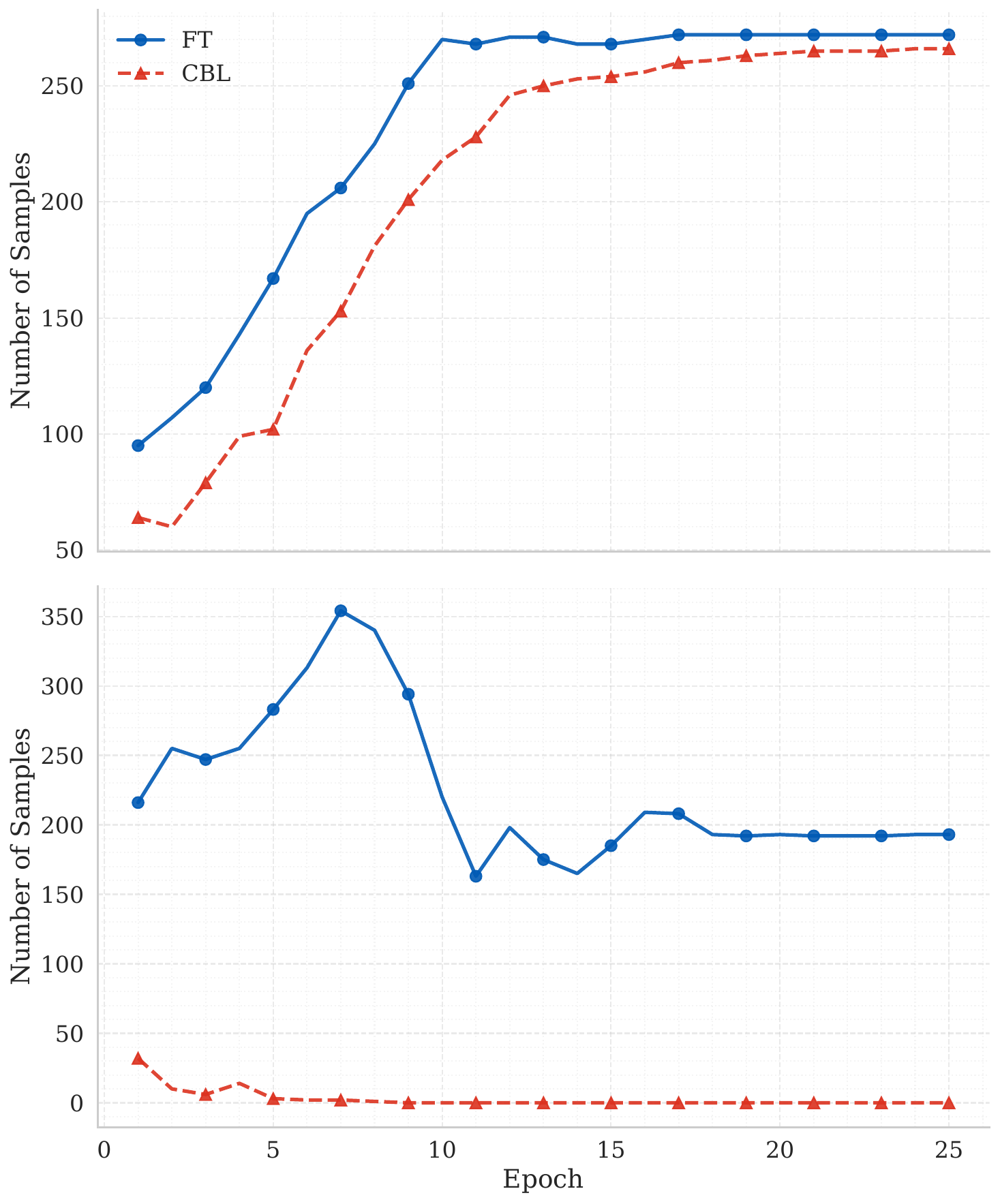}
\end{subfigure}

\caption{Learning and forgetting curves across four datasets and four optimizers. The rows represent SGD, Momentum, Adam, and AdamW. The inner labels indicate the metric type (Learning vs Forgetting) for the corresponding vertical position.}
\label{fig:full_optimizer_comparison}
\end{figure*}

The dynamic characteristics of learning and forgetting during knowledge injection are shown in Fig.~\ref{fig:full_optimizer_comparison}. Although CPL freezes Conflicting Parameters and only updates Collaborative Parameters, the speed of knowledge injection for CPL is not noticeably slowed than FT, This indicates that Collaborative Parameters are sufficient for training, and the knowledge injection capability of CPL is not compromised by freezing Conflicting Parameters.

\section{Extending CPL to More Optimizers}
\label{app:theoretical}

\subsection{Momentum Optimizer}
When performing gradient descent with learning rate $\eta$, the update rule for parameters $\theta$ in the Momentum optimizer is as follows:
\begin{equation}
\begin{aligned}
    m &\leftarrow \beta m + \nabla_\theta \mathcal{L}^\mathcal{I}(\theta) \\
    \theta &\leftarrow \theta - \eta m
\end{aligned}
\end{equation}
where $m$ performs an exponential weighted average of historical gradients for accelerated convergence. Similar to Eq.~\eqref{eq:loss_exp}, the first-order change in loss on Mastered Set is
\begin{equation}
    \Delta \mathcal{L}^\mathcal{M} \approx -\eta \underbrace{\nabla_\theta \mathcal{L}^\mathcal{M}(\theta)^\top m}_{S^{mom}(\mathcal{M}, \mathcal{I})}.
\end{equation}
Here, $S^{mom}(\mathcal{M}, \mathcal{I})$ is defined as the momentum similarity. Utilizing the additivity of inner product, we decompose the global momentum similarity onto each neuron:
\begin{equation}
    S^{mom}(\mathcal{M}, \mathcal{I}) = \sum_{j=1}^{|\theta|} s_{\theta_j}^{mom},
    \quad
    s_{\theta_j}^{mom} = \nabla_{\theta_j} \mathcal{L}^\mathcal{M}(\theta) \cdot m_j.
\end{equation}
The parameter update rule of CPL for momentum is
\begin{equation}
    \label{eq:momentum_update}
    \theta \leftarrow \theta - \eta \left[ \mathbb{I}(S^{mom} \geq 0) \odot m \right].
\end{equation}
Under this rule, the first-order loss change on Mastered Set is
\begin{equation}
\begin{aligned}
    \Delta \mathcal{L}^\mathcal{M}
    &\approx -\eta \sum_{j=1}^{|\theta|} I(s_{\theta_j}^{mom} \geq 0) \cdot s_{\theta_j}^{mom}
    \leq 0.
\end{aligned}
\end{equation}

\subsection{Adam Optimizer (without weight decay)}
The conventional update rule for the Adam optimizer is as follows:
\begin{equation}
\begin{aligned}
    m &\leftarrow \beta_1 m + (1 - \beta_1) g_t \\
    v &\leftarrow \beta_2 v + (1 - \beta_2) g_t^2 \\
    \hat{m} &= \frac{m}{1 - \beta_1^t}, \quad \hat{v} = \frac{v}{1 - \beta_2^t} \\
    \theta &\leftarrow \theta - \eta \frac{\hat{m}}{\sqrt{\hat{v}} + \epsilon}
\end{aligned}
\end{equation}
where $g_t = \nabla_\theta \mathcal{L}^\mathcal{I}(\theta)$. Denote the actual Adam update direction as
\begin{equation}
    U^{adam} = \frac{\hat{m}}{\sqrt{\hat{v}} + \epsilon}.
\end{equation}
The first-order change in loss on Mastered Set is
\begin{equation}
    \Delta \mathcal{L}^\mathcal{M} \approx -\eta \underbrace{\nabla_\theta \mathcal{L}^\mathcal{M}(\theta)^\top U^{adam}}_{S^{adam}(\mathcal{M}, \mathcal{I})}.
\end{equation}
We decompose the global Adam similarity as
\begin{equation}
    S^{adam}(\mathcal{M}, \mathcal{I}) = \sum_{j=1}^{|\theta|} s_{\theta_j}^{adam},
    \quad
    s_{\theta_j}^{adam} = \nabla_{\theta_j} \mathcal{L}^\mathcal{M}(\theta) \cdot U^{adam}_j.
\end{equation}
The CPL update rule for Adam is
\begin{equation}
    \label{eq:adam_update}
    \theta \leftarrow \theta - \eta \left[ \mathbb{I}(S^{adam} \geq 0) \odot U^{adam} \right],
\end{equation}
which yields
\begin{equation}
\begin{aligned}
    \Delta \mathcal{L}^\mathcal{M}
    &\approx -\eta \sum_{j=1}^{|\theta|} I(s_{\theta_j}^{adam} \geq 0) \cdot s_{\theta_j}^{adam}
    \leq 0.
\end{aligned}
\end{equation}

\subsection{AdamW Optimizer}
The AdamW optimizer introduces decoupled weight decay on top of Adam:
\begin{equation}
\begin{aligned}
    m &\leftarrow \beta_1 m + (1 - \beta_1) g_t \\
    v &\leftarrow \beta_2 v + (1 - \beta_2) g_t^2 \\
    \hat{m} &= \frac{m}{1 - \beta_1^t}, \quad \hat{v} = \frac{v}{1 - \beta_2^t} \\
    \theta &\leftarrow \theta - \eta \left( \frac{\hat{m}}{\sqrt{\hat{v}} + \epsilon} + \lambda \theta \right).
\end{aligned}
\end{equation}
For notational convenience, we denote the actual total update direction calculated by AdamW as
\begin{equation}
    U^{adamw} = \frac{\hat{m}}{\sqrt{\hat{v}} + \epsilon} + \lambda \theta.
\end{equation}
The first-order loss change is
\begin{equation}
    \Delta \mathcal{L}^\mathcal{M} \approx -\eta \underbrace{\nabla_\theta \mathcal{L}^\mathcal{M}(\theta)^\top U^{adamw}}_{S^{adamw}(\mathcal{M}, \mathcal{I})}.
\end{equation}
We decompose it as
\begin{equation}
    S^{adamw}(\mathcal{M}, \mathcal{I}) = \sum_{j=1}^{|\theta|} s_{\theta_j}^{adamw},
    \quad
    s_{\theta_j}^{adamw} = \nabla_{\theta_j} \mathcal{L}^\mathcal{M}(\theta) \cdot U^{adamw}_j.
\end{equation}
The CPL update rule for AdamW is
\begin{equation}
    \theta \leftarrow \theta - \eta \left[ \mathbb{I}(S^{adamw} \geq 0) \odot U^{adamw} \right],
\end{equation}
and the first-order loss change satisfies
\begin{equation}
\begin{aligned}
    \Delta \mathcal{L}^\mathcal{M}
    &\approx -\eta \sum_{j=1}^{|\theta|} I(s_{\theta_j}^{adamw} \geq 0) \cdot s_{\theta_j}^{adamw}
    \leq 0.
\end{aligned}
\end{equation}

\subsection{Arbitrary Optimizers}
\label{app:general_optimizers}
To derive a unified first-order framework, we abstract the specific update rules of different optimizers into a generalized form. Regardless of whether the optimizer utilizes momentum, adaptive learning rates, or weight decay, its parameter update can ultimately be expressed as applying a specific update vector $F^{opt}(\nabla_\theta \mathcal{L}^\mathcal{I}(\theta))$ to the current parameters:
\begin{equation}
    \theta \leftarrow \theta - \eta F^{opt}(\nabla_\theta \mathcal{L}^\mathcal{I}(\theta)).
\end{equation}
The first-order loss change on the mastered set is
\begin{equation}
    \Delta \mathcal{L}^\mathcal{M} \approx -\eta \underbrace{\nabla_\theta \mathcal{L}^\mathcal{M}(\theta)^\top F^{opt}(\nabla_\theta \mathcal{L}^\mathcal{I}(\theta))}_{S^{opt}(\mathcal{M}, \mathcal{I})}.
\end{equation}
Utilizing the additivity of the inner product, we decompose the global similarity onto each neuron:
\begin{equation}
    S^{opt}(\mathcal{M}, \mathcal{I}) = \sum_{j=1}^{|\theta|} s_{\theta_j}^{opt},
    \quad
    s_{\theta_j}^{opt} = \nabla_{\theta_j} \mathcal{L}^\mathcal{M}(\theta) \cdot F^{opt}_j(\nabla_\theta \mathcal{L}^\mathcal{I}(\theta)).
\end{equation}
The generalized CPL parameter update formula is
\begin{equation}
    \label{eq:general_update}
    \theta \leftarrow \theta - \eta \left[ \mathbb{I}(S^{opt} \geq 0) \odot F^{opt}(\nabla_\theta \mathcal{L}^\mathcal{I}(\theta)) \right].
\end{equation}
Substituting the generalized CPL update rule back into the first-order approximation gives
\begin{equation}
\begin{aligned}
    \Delta \mathcal{L}^\mathcal{M}
    &\approx -\eta \sum_{j=1}^{|\theta|} I(s_{\theta_j}^{opt} \geq 0) \cdot s_{\theta_j}^{opt}
    \leq 0.
\end{aligned}
\end{equation}

\subsection{Experimental Results}

When applying CPL to Momentum, Adam, and AdamW optimizers, we observe consistent performance in mitigating catastrophic forgetting across all four datasets. As shown in Table~\ref{tab:optimizers}, regardless of the optimizer used, CPL achieves zero forgetting while maintaining effective knowledge injection, demonstrating its general applicability across different optimization.

\begin{table}[h]
\centering
\caption{Comparison of knowledge injection performance across different optimizers. \textit{Qwen2.5 1.5B} is selected for experiment. Whether using SGD, Momentum, Adam, or AdamW, CPL consistently forgets zero question when Mastered Set $\mathcal{M}$ is fully accessed.}
\label{tab:optimizers}
\begin{small}
\begin{sc}
\resizebox{\textwidth}{!}{
\begin{tabular}{llcccccccc}
\toprule
\multirow{2}{*}{\textbf{Optimizer}} & \multirow{2}{*}{\textbf{Method}} & \multicolumn{2}{c}{\textbf{MMLU}} & \multicolumn{2}{c}{\textbf{MedQA}} & \multicolumn{2}{c}{\textbf{ARC-C}} & \multicolumn{2}{c}{\textbf{CSQA}} \\
\cmidrule(lr){3-4} \cmidrule(lr){5-6} \cmidrule(lr){7-8} \cmidrule(lr){9-10}
 & & \textbf{Learned} & \textbf{Forgot} & \textbf{Learned} & \textbf{Forgot} & \textbf{Learned} & \textbf{Forgot} & \textbf{Learned} & \textbf{Forgot} \\
\midrule
\multirow{4}{*}{SGD} 
 & \multirow{2}{*}{FT} & 168 & 188 & 191 & 264 & 104 & 88 & 86 & 163 \\
 & & (24.8\%) & (21.4\%) & (27.1\%) & (46.5\%) & (32.1\%) & (9.3\%) & (31.6\%) & (19.8\%) \\
\cmidrule(lr){2-10}
 & \multirow{2}{*}{CPL} & 168 & \textbf{0} & 193 & \textbf{0} & 163 & \textbf{0} & 147 & \textbf{0} \\
 & & (24.8\%) & (\textbf{0.0\%}) & (27.1\%) & (\textbf{0.0\%}) & (50.3\%) & (\textbf{0.0\%}) & (54.0\%) & (\textbf{0.0\%}) \\
\midrule
\multirow{4}{*}{Momentum} 
 & \multirow{2}{*}{FT} & 341 & 450 & 257 & 398 & 261 & 169 & 205 & 233 \\
 & & (50.3\%) & (51.3\%) & (36.5\%) & (70.1\%) & (80.6\%) & (17.9\%) & (75.4\%) & (28.3\%) \\
\cmidrule(lr){2-10}
 & \multirow{2}{*}{CPL} & 408 & \textbf{0} & 461 & \textbf{0} & 291 & \textbf{0} & 240 & \textbf{0} \\
 & & (60.2\%) & (\textbf{0.0\%}) & (65.5\%) & (\textbf{0.0\%}) & (89.8\%) & (\textbf{0.0\%}) & (88.2\%) & (\textbf{0.0\%}) \\
\midrule
\multirow{4}{*}{Adam} 
 & \multirow{2}{*}{FT} & 647 & 203 & 704 & 181 & 320 & 143 & 266 & 168 \\
 & & (95.4\%) & (23.1\%) & (100.0\%) & (31.9\%) & (98.8\%) & (15.1\%) & (97.8\%) & (20.4\%) \\
\cmidrule(lr){2-10}
 & \multirow{2}{*}{CPL} & 648 & \textbf{0} & 704 & \textbf{0} & 323 & \textbf{0} & 266 & \textbf{0} \\
 & & (95.6\%) & (\textbf{0.0\%}) & (100.0\%) & (\textbf{0.0\%}) & (99.7\%) & (\textbf{0.0\%}) & (97.8\%) & (\textbf{0.0\%}) \\
\midrule
\multirow{4}{*}{AdamW} 
 & \multirow{2}{*}{FT} & 638 & 195 & 703 & 174 & 320 & 143 & 251 & 294 \\
 & & (94.1\%) & (22.2\%) & (99.9\%) & (30.6\%) & (98.8\%) & (15.1\%) & (92.3\%) & (35.7\%) \\
\cmidrule(lr){2-10}
 & \multirow{2}{*}{CPL} & 642 & \textbf{0} & 703 & \textbf{0} & 323 & \textbf{0} & 266 & \textbf{0} \\
 & & (94.7\%) & (\textbf{0.0\%}) & (99.9\%) & (\textbf{0.0\%}) & (99.7\%) & (\textbf{0.0\%}) & (97.8\%) & (\textbf{0.0\%}) \\
\bottomrule
\end{tabular}
}
\end{sc}
\end{small}
\vskip -0.1in
\end{table}

\section{Extending CPL to Various Tasks}
\label{app:continual_learning}

CPL is extended to various tasks including: sentiment classification (Yahoo Answers, AG News, and DBPedia) \cite{zhang2015character}, open-ended question answering (NumGLUE-cm and NumGLUE-ds) \cite{mishra2022numglue}, multilingual settings (C-STANCE) \cite{zhao2023cstance}, and multimodal tasks (MMBench) \cite{liu2024mmbench}. In Table~\ref{tab:twelve_task_order1}, each cell is reported as ``Learned/Forgotten''. CPL keeps the forgotten count much smaller across both tasks compared with FT, demonstrating its effectiveness across a wide range of tasks.

\begin{table*}[h]
\centering
\caption{Learning results on 7 tasks. Each cell is recorded as ``Learned/Forgotten''. CPL keeps the forgotten count much smaller across both tasks compared with FT.}
\label{tab:twelve_task_order1}
\begin{small}
\begin{sc}
\resizebox{\textwidth}{!}{
\begin{tabular}{lccccccc}
\toprule
\textbf{Dataset} & \textbf{Yahoo} & \textbf{AG} & \textbf{DBpedia} & \textbf{Num-ds} & \textbf{Num-cm} & \textbf{C-STANCE} & \textbf{MMBench} \\
\midrule
FT & 50/102 & 25/10 & 29/5 & 71/14 & 54/41 & 13/13 & 115/60 \\
\textbf{CPL} & \textbf{50/0} & \textbf{30/1} & \textbf{29/0} & \textbf{87/5} & \textbf{57/7} & \textbf{21/0} & \textbf{116/0} \\
\bottomrule
\end{tabular}
}
\end{sc}
\end{small}
\vskip -0.1in
\end{table*}

\section{Extending CPL to Out-of-Set Generalization}
\label{app:out_of_set_generalization}

The previous experiments focus on the ideal scenario where the Mastered Set $\mathcal{M}$ is fully accessible. To evaluate the out-of-set generalization capability of CPL, we design an experiment where both Injection Set and Mastered Set are partitioned into training and evaluation subsets, respectively. The mask is computed from the training subset, while the evaluation subset is used to assess learning and forgetting performance.

In Table~\ref{tab:held_out_eval}, ``Injection'' refers to splitting the Injection Set into training and evaluation subsets, while the Mastered Set remains fully accessible for mask computation. Since the Mastered Set is available, CPL still forgets zero question. However, due to distribution shifts between the training and evaluation subsets of the Injection Set, the number of learned questions is reduced. However, CPL still learn comparable number of questions compared with FT.

``Mastered'' refers to splitting the Mastered Set into training and evaluation subsets, while the Injection Set remains fully accessible for training. Since the gradient distribution of evaluation subset can only be partially estimated from the training subset, CPL forgets more questions compared with the ideal scenario. Nevertheless, CPL still reduces forgetting by 59.1\%–81.7\% compared with FT, and incorporating the training subset of the Mastered Set using CPL is more effective than using replay.

\begin{table*}[h]
\centering
\caption{Performance evaluation under out-of-set setting. \textit{Qwen2.5 1.5B} is selected for experiment. To assess generalization, Injection Set and Mastered Set are partitioned into training and evaluation subsets, respectively. This table reports the learning and forgetting performance on unseen evaluation data, demonstrating the out-of-set generalization capability of CPL.}
\label{tab:held_out_eval}
\begin{small}
\begin{sc}
\resizebox{\textwidth}{!}{
\begin{tabular}{llcccccccc}
\toprule
 & \multirow{2}{*}{\textbf{Method}} & \multicolumn{2}{c}{\textbf{MMLU}} & \multicolumn{2}{c}{\textbf{MedQA}} & \multicolumn{2}{c}{\textbf{ARC-C}} & \multicolumn{2}{c}{\textbf{CSQA}} \\
\cmidrule(lr){3-4} \cmidrule(lr){5-6} \cmidrule(lr){7-8} \cmidrule(lr){9-10}
 & & \textbf{Learned} & \textbf{Forgot} & \textbf{Learned} & \textbf{Forgot} & \textbf{Learned} & \textbf{Forgot} & \textbf{Learned} & \textbf{Forgot} \\
\midrule
\multirow{4}{*}{Injection} 
 & \multirow{2}{*}{FT}  & 24 & 48 & 34 & 75 & 29 & 71 & 22 & 114 \\
 &                      & (7.3\%) & (5.5\%) & (9.7\%) & (13.2\%) & (17.9\%) & (7.5\%) & (16.2\%) & (13.8\%) \\
\cmidrule{2-10}
 & \multirow{2}{*}{CPL} & 31 & \textbf{0} & 39 & \textbf{0} & 29 & \textbf{0} & 23 & \textbf{0} \\
 &                      & (9.5\%) & (\textbf{0.0\%}) & (11.1\%) & (\textbf{0.0\%}) & (17.9\%) & (\textbf{0.0\%}) & (16.9\%) & (\textbf{0.0\%}) \\
\midrule
\multirow{6}{*}{Mastered} 
 & \multirow{2}{*}{FT}  & 178 & 110 & 208 & 143 & 104 & 49 & 86 & 94 \\
 &                      & (26.3\%) & (26.0\%) & (29.5\%) & (50.4\%) & (32.1\%) & (10.4\%) & (31.6\%) & (22.8\%) \\
\cmidrule{2-10}
 & \multirow{2}{*}{Replay}  & 157 & 61 & 124 & 58 & 86 & 14 & 70 & 28 \\
 &                      & (23.2\%) & (14.4\%) & (17.6\%) & (20.4\%) & (26.5\%) & (3.0\%) & (25.7\%) & (6.8\%) \\
\cmidrule{2-10}
 & \multirow{2}{*}{CPL} & 179 & \textbf{46} & 243 & \textbf{33} & 116 & \textbf{9} & 147 & \textbf{18} \\
 &                      & (26.4\%) & (\textbf{10.9\%}) & (34.5\%) & (\textbf{11.6\%}) & (35.8\%) & (\textbf{1.9\%}) & (54.0\%) & (\textbf{4.4\%}) \\
\bottomrule
\end{tabular}
}
\end{sc}
\end{small}
\vskip -0.1in
\end{table*}

\section{Extending CPL to Cross-Prompt Generalization}
\label{app:cross_prompt}

To evaluate CPL inject the specific prompt format or the underlying knowledge, cross-prompt generalization is studided. Four prompts as following are designed for each question. A question enters the Mastered Set only if all prompts are answered correctly, and enters the Injection Set only if all prompts are answered incorrectly. The first and third prompts are used for training, while the second and fourth prompts are used for evaluation.

Table~\ref{tab:cross_prompt} reports CPL mitigates forgetting under unseen prompt formats while maintaining comparable learning performance. 

\noindent\textbf{template1}
\begin{verbatim}
Output exactly one uppercase option label.

Question:
{question}

Options:
A. {option_a}
B. {option_b}
C. {option_c}
D. {option_d}

Use only A, B, C, or D. Do not explain your answer.
\end{verbatim}

\noindent\textbf{template2}
\begin{verbatim}
Choose the best answer to the question below.

Question:
{question}

Selections:
A. {option_a}
B. {option_b}
C. {option_c}
D. {option_d}

Reply with a single capital letter: A, B, C, or D.
Do not add any explanation.
\end{verbatim}

\noindent\textbf{template3}
\begin{verbatim}
Read the problem below.

{question}

Output exactly one number: 1, 2, 3, or 4.
Choices:
1) {option_a}
2) {option_b}
3) {option_c}
4) {option_d}

No reasoning is required.
\end{verbatim}

\noindent\textbf{template4}
\begin{verbatim}
Consider the following question.

Question:
{question}

Answer set:
1) {option_a}
2) {option_b}
3) {option_c}
4) {option_d}

Respond with one digit only: 1, 2, 3, or 4.
No explanation should appear in your response.
\end{verbatim}

\begin{table*}[h]
\centering
\caption{Cross-prompt performance of CPL. CPL substantially reduces forgetting under unseen prompt formats while maintaining comparable learning.}
\label{tab:cross_prompt}
\begin{small}
\begin{sc}
\resizebox{\textwidth}{!}{
\begin{tabular}{lcccccccc}
\toprule
\multirow{2}{*}{\textbf{Method}} & \multicolumn{2}{c}{\textbf{MMLU}} & \multicolumn{2}{c}{\textbf{MedQA}} & \multicolumn{2}{c}{\textbf{ARC-C}} & \multicolumn{2}{c}{\textbf{CSQA}} \\
\cmidrule(lr){2-3} \cmidrule(lr){4-5} \cmidrule(lr){6-7} \cmidrule(lr){8-9}
 & \textbf{Learned} & \textbf{Forgot} & \textbf{Learned} & \textbf{Forgot} & \textbf{Learned} & \textbf{Forgot} & \textbf{Learned} & \textbf{Forgot} \\
\midrule
FT & 24 & 263 & 47 & 207 & 22 & 375 & 28 & 321 \\
CPL & 27 & \textbf{98} & 48 & \textbf{76} & 24 & \textbf{88} & 28 & \textbf{75} \\
\bottomrule
\end{tabular}
}
\end{sc}
\end{small}
\vskip -0.1in
\end{table*}



\end{document}